\documentclass[journal]{IEEEtran}
\usepackage{amsmath,amsfonts}
\usepackage{algorithmic}
\usepackage{algorithm}
\usepackage{array}
\usepackage[caption=false,font=scriptsize,labelfont=sf,textfont=sf]{subfig}
\usepackage{textcomp}
\usepackage{stfloats}
\usepackage{url}
\usepackage{verbatim}
\usepackage{graphicx}
\usepackage{cite}
\usepackage[acronym]{glossaries}
\usepackage{booktabs}
\usepackage{multirow}
\usepackage{tikz}
\usepackage{makecell}

\begin{document}

\title{Regulating Autonomous Mobility-on-Demand Operators with Fixed and Adaptive Penalties: A Multi-Agent Reinforcement Learning Approach}

\author{Korbinian Kügler$^{1}$, Carolin Schmidt$^{1}$, Florian Dandl$^{1}$, Filipe Rodrigues$^{2}$

 \thanks{$^{1}$Technical University of Munich, Germany 
 		{\tt\small \{korbinian.kuegler, carolin.schmidt, florian.dandl\}@tum.de}}%
\thanks{$^{2}$Technical University of Denmark, Denmark
 		{\tt\small rodr@dtu.dk}}
}


\maketitle

\begin{abstract}
    Profit-maximizing Autonomous Mobility-on-Demand (AMoD) operators may exploit consumer surplus through excessive prices and underserve less profitable regions by relocating vehicles away from them. We study how penalty-based regulation of an AMoD operator in a multimodal mobility system can limit surge pricing and improve regional service availability. We model regulation as an outer control loop that acts on a learning-based operator. Static regulation with fixed penalties is formulated as a Markov Decision Process for an operator that jointly controls pricing and rebalancing, and dynamic regulation as a Partially Observable Stochastic Game in which a regulator periodically adjusts region-specific penalties in closed loop. Operator and regulator policies are learned with graph-based multi-agent reinforcement learning. In experiments with real-world Manhattan data, fixed surge pricing penalties reduce the share of travelers facing excessive prices from 22\% to 1\%. For regional service availability, dynamic regulation improves the regulator's reward by up to 15.6\% over fixed penalties of equal mean intensity. The gains in service availability are strongest when spare fleet capacity exists, whereas service deficits persist in peak periods with highly utilized fleets. The learned regulator policy responds by reducing penalties where limited fleet capacity restricts compliance. The results identify trade-offs between regulatory targets, operator profitability, and traveler utility. 
\end{abstract}



\newacronym{acr:amod}{AMoD}{autonomous Mobility-on-Demand}
\newacronym{acr:mod}{MoD}{Mobility-on-Demand}
\newacronym{acr:pt}{PT}{public transport}
\newacronym{acr:msp}{MSP}{mobility service provider}
\newacronym[longplural={municipal transportation authorities}]{acr:mta}{MTA}{municipal transportation authority}

\newacronym{acr:mpc}{MPC}{Model Predictive Control}
\newacronym{acr:ppo}{PPO}{Proximal Policy Optimization}

\newacronym{acr:mdp}{MDP}{Markov Decision Process}
\newacronym{acr:posg}{POSG}{Partially Observable Stochastic Game}

\newacronym{acr:drl}{DRL}{deep reinforcement learning}
\newacronym{acr:rl}{RL}{reinforcement learning}

\newacronym{acr:sac}{SAC}{Soft Actor-Critic}
\newacronym{acr:a2c}{A2C}{Advantage Actor-Critic}
\newacronym{acr:td}{TD}{temporal-difference}

\newacronym{acr:gcn}{GCN}{graph convolutional network}
\newacronym{acr:fc}{FC}{fully connected}
\newacronym{acr:lrelu}{LeakyReLU}{Leaky Rectified Linear Unit}
\newacronym{acr:relu}{ReLU}{Rectified Linear Unit}

\newacronym{acr:fcfs}{FCFS}{first-come, first-served}
\newacronym{acr:od}{OD}{origin-destination}

\newacronym{acr:gtfs}{GTFS}{General Transit Feed Specification}


\section{Introduction} \label{chap:intro}
\IEEEPARstart{R}{ising} mobility demand puts pressure on urban transportation systems \cite{itf2023transportoutlook}, motivating \gls{acr:mod} services aimed at utilizing existing infrastructure more efficiently \cite{erhardt2019congestion}. \Gls{acr:amod} represents the next evolutionary step of these services and is moving from concept to operational reality. 
By replacing human-driven vehicles with fully compliant autonomous vehicles under centralized fleet control, \gls{acr:amod} operators can steer supply by continuously rebalancing idle vehicles without relying on driver participation \cite{salazar2018interactionamodpt} and influence demand through dynamic pricing without accounting for driver-side incentives \cite{lei2019dynamicpricing}. 
However, profit-oriented operators may use centralized control to concentrate service in high-revenue regions \cite{gao2024regulatingforhireav ,Syed.2021b
} and capture consumer surplus through price discrimination \cite{bahamonde-birke2021necessity,lanzetti2024interplayamodandpt}, motivating regulation that aligns \gls{acr:amod} operations with public-sector objectives \cite{bahamonde-birke2018policyaspects}.
 
Designing such regulatory frameworks is complex, as regulators must balance citizen satisfaction, operator interests, and sustainability goals \cite{zardini2021gametheory}. Interventions may also be limited by the temporal mismatch between operators' frequent operational decisions \cite{li2025joint} and medium- to long-term regulatory adjustments \cite{dandl2021regulating}. At the same time, requirements for sharing dynamic mobility data \cite{mdv2022anlage} together with the algorithmic controllability of \gls{acr:amod} fleets make adaptive regulation a practically relevant alternative to fixed instruments. 

Still, it remains unclear how regulation affects an \gls{acr:amod} operator whose decisions adapt to the system state. Existing work either learns operator policies without external incentives or treats regulation as fixed instruments evaluated against strategic or steady-state operator responses. Neither the effect of fixed regulation on a learned operator nor incentives imposed on the operator that adapt to its behavior in closed loop have been examined. Recent work \cite{kadem2026dynamicincentivization} adapts \gls{acr:pt} fare reductions rather than incentives imposed on the operator.

We introduce the \textit{\gls{acr:amod} regulation and control problem}, in which penalties, targeted on regional service availability and excessive surge pricing, influence a profit-maximizing operator while travelers choose modes. 

Our contributions are threefold: 
\begin{itemize}
    \item We formulate the \gls{acr:amod} regulation and control problem for static regulation as a \gls{acr:mdp}, in which the operator decides under fixed penalties, and for dynamic regulation as a \gls{acr:posg}, in which a partially observing regulator adjusts penalties in closed loop at a lower decision frequency and pursues objectives distinct from the operator's.
    \item We develop a multi-agent \gls{acr:rl} framework that learns the regulator's policy jointly with the pricing and rebalancing policy of a profit-maximizing operator, enabling penalties that adapt to observed operational data.
    \item In a Manhattan case study with real-world data, we show that fixed penalties suffice against excessive surge pricing, whereas for regional service availability, adaptive regulation outperforms fixed penalties of equal mean intensity by up to 15.6\% in regulator reward  and reduces penalties where limited fleet capacity restricts compliance. We further analyze operator adaptation and identify trade-offs among regulatory targets, operator profitability, and traveler utility.
\end{itemize}


\section {Literature Review}  \label{chap:litrev}
We connect the streams on operational control of \gls{acr:amod} systems and on regulation in multimodal \gls{acr:amod} systems.

\subsection{Operational Control of AMoD Systems}
Early rebalancing approaches use queueing-theoretical models \cite{zhang2015control}, model predictive control \cite{iglesias2018mpc}, and robust optimization \cite{guo2021uncertaindemand}. \Gls{acr:rl} provides an alternative to repeatedly solving model-based optimization problems online. Gammelli et al. \cite{gammelli2021gnnrl} propose a bi-level framework combining optimization and \gls{acr:rl} based on \gls{acr:gcn} architectures exploiting transportation-network structure. 

Dynamic pricing complements rebalancing by shaping demand. Joint optimization can improve operator profits relative to separate control \cite{wollenstein2020joint}. Li et al. \cite{li2025joint} develop a graph-based \gls{acr:rl} framework for joint pricing and rebalancing, extended by Toft et al. \cite{toft2026competitive} to competing operators and price-sensitive passenger choice. These approaches provide the methodological basis for operator control but do not consider external regulatory objectives and incentives. Although Bala et al. \cite{bala2026equitableoperation} incorporate equity into the operator's reward, equity is an internal objective rather than an outcome of regulation imposed on a self-interested operator.

\subsection{Regulation in Multimodal AMoD Systems} \label{sec:litrev_regulation}
Research on regulation in multimodal \gls{acr:amod} systems examines the influence of interventions on profit-oriented operators and mobility outcomes. A first group of studies evaluates exogenously specified regulatory instruments in game-theoretic frameworks, including \gls{acr:amod} service taxes and fleet-size restrictions \cite{lanzetti2024interplayamodandpt}, as well as minimum service requirements and tax-funded \gls{acr:pt} subsidies \cite{gao2024regulatingforhireav}.

Other studies explicitly determine interventions through Bayesian optimization \cite{dandl2021regulating}, backward induction \cite{zardini2021gametheory,zardini2023strategicinteractions}, feedback optimization \cite{he2025layeredmobility}, and implementation-based incentive design \cite{li2026incentivedesign}. However, they generally represent operator responses through service-design choices or steady-state operating plans, with regulation optimized as a medium- to long-term policy choice and applied open-loop, i.e., held fixed while operators and travelers respond.

Moving toward dynamic, closed-loop operational interaction, Kadem et al. \cite{kadem2026dynamicincentivization} propose a multi-agent \gls{acr:rl} framework in which a regulator dynamically reduces \gls{acr:pt} fares and a profit-maximizing mobility service provider prices trips. However, the intervention targets \gls{acr:pt} users, and the learned operator policy controls pricing without jointly optimizing rebalancing. We address the gap at the intersection of the two streams by studying the effectiveness of fixed (open-loop) and adaptive (closed-loop) penalties imposed on a profit-maximizing \gls{acr:amod} operator who jointly controls prices and vehicle rebalancing.


\section{Problem Setting} \label{chap:problem}
We consider a multimodal urban mobility system with a \textit{regulator}, an \gls{acr:amod} \textit{operator}, \textit{\gls{acr:pt}}, and \textit{travelers}.
The regulator influences the operator through monetary penalties for deviations from public-sector targets without directly prescribing operational decisions. The operator sets \gls{acr:od}-based prices and rebalances idle vehicles to maximize revenues minus operating costs and penalties. \Gls{acr:pt} serves as an alternative mode, including bus and subway connections with walking access and egress. Its fares, routes, and frequencies remain fixed and exogenous, reflecting the longer service-design horizon \cite{zardini2021gametheory}. Travelers choose between \gls{acr:amod} and \gls{acr:pt} based on travel times and monetary costs.
 
We consider two penalty-based instruments, applied separately, that address two levers through which a profit-maximizing operator can lower its level of service, namely charging excessive prices and withdrawing vehicles from less profitable regions.
\textit{Regional service availability regulation} penalizes regional service rates below a target, where the service rate is the share of regional traveler requests served by \gls{acr:amod}. \textit{Surge pricing regulation} penalizes offered prices above a threshold relative to historical fares. We assume that each penalty is proportional to the corresponding deficit or excess and the penalty level set by the regulator. 

Under \textit{static regulation}, exogenous penalty levels are fixed and spatially uniform. Under \textit{dynamic regulation}, the regulator periodically adjusts region-specific penalty levels based on observations to optimize its objective. It acts less frequently than the operator, reducing sensitivity to short-term demand noise and avoiding rapid changes in operator incentives. We evaluate both instruments under static regulation but restrict dynamic regulation to regional service availability, as static surge pricing regulation nearly eliminates excessive prices with negligible penalty payments (see Sec.~\ref{sec:results_surge_pricing_performance}).


\section{Methodology} \label{chap:methodology}
We represent the transportation system as a directed graph $\mathcal{G} = (\mathcal{V}, \mathcal{E})$, whose nodes are regions with central pick-up and drop-off stations and whose edges represent shortest paths between them. Time evolves in finite discrete steps $t\in\mathcal{T}=\{0,\dots,H-1\}$ with duration $\Delta^{\mathrm{AMOD}}$ over horizon $H$. Travel between stations $i$ and $j$ by mode $m\in\mathcal{M}=\{\mathrm{AMOD},\mathrm{PT}\}$ has time-dependent duration $\tau_{i,j,t}^{m}\in\mathbb{Z}_+$ and price $p_{i,j,t}^{m}$. The operator controls $N^{\mathrm{AMOD}}$ single-passenger vehicles. Vehicles may either serve passengers or be rebalanced, both incurring costs $c_{i,j,t}^{\mathrm{AMOD}}$. Binary logit mode choice partitions traveler requests $d_{i,j,t}$ into $d_{i,j,t}^{\mathrm{AMOD}}$ and $d_{i,j,t}^{\mathrm{PT}}$. Travelers choosing \gls{acr:amod} queue at their origin, are matched to local vehicles on a \gls{acr:fcfs} basis, and will switch to \gls{acr:pt} if their waiting time exceeds a threshold $\omega$.

\subsection{Decision Architecture}
\begin{figure*}[t]
    \centering
    \includegraphics[width=\linewidth]{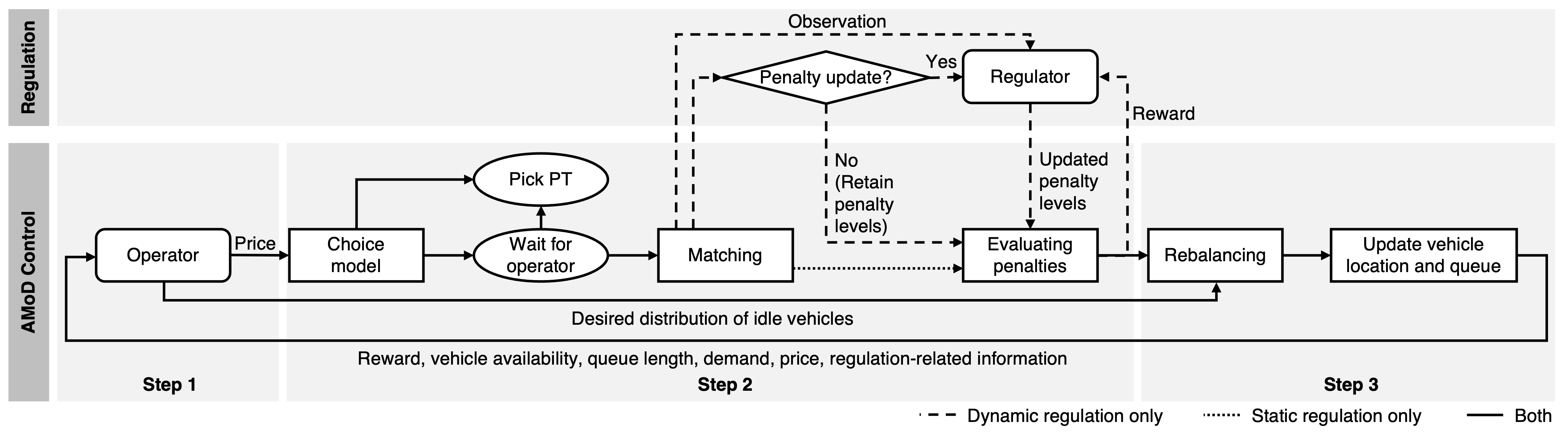}
    \caption[Decision architecture]{Decision architecture, consisting of an \gls{acr:amod} control layer and, under dynamic regulation, an \gls{acr:amod} regulation layer. In the \gls{acr:amod} control layer, Step 1 determines prices and target idle-vehicle distribution, Step 2 captures mode choice, queueing, passenger matching, and regulation evaluation, and Step 3 executes rebalancing and updates vehicle locations and queues. In the \gls{acr:amod} regulation layer, penalty levels are updated at each regulatory decision interval. Between regulatory updates, the current penalty levels remain in force.}
    \label{fig:decision_architecture}
\end{figure*}

The decision architecture has two interacting layers for \gls{acr:amod} control and regulation (Fig.~\ref{fig:decision_architecture}). The operator follows the bi-level formulation of \cite{gammelli2023grlbi-level,li2025joint} adapted to a regulated setting. Each time step has three stages: \textit{(1)} the operator policy sets origin-based price factors and desired idle-vehicle shares; \textit{(2)} stochastic demand is generated, travelers choose modes, queued \gls{acr:amod} requests are matched, wait, or leave the \gls{acr:amod} service for \gls{acr:pt}, and penalties are evaluated; \textit{(3)} minimum-cost optimization converts desired idle-vehicle shares into rebalancing flows (see Appendix), after which vehicles and queues are updated and the next state and reward returned. Under dynamic regulation, the regulator updates penalties every $\Delta^{\mathrm{REG}}$ operator steps.

\subsection{AMoD Regulation and Control Problem as MDP and POSG} \label{sec:mdp}
\subsubsection{Static Regulation as \gls{acr:mdp}}
Static regulation yields a single-agent \gls{acr:mdp} $(\mathcal{S},\mathcal{A},R,T)$.

\textit{State ($\mathcal{S}$):}
The state $\mathbf{s}_t$ at $t \in \mathcal{T}$ comprises: the network adjacency matrix $\mathbf{A}$, current time step $t$, and for each region $i \in \mathcal{V}$: current idle vehicles $n_{i,t}^{\mathrm{AMOD}} \in [0, N^{\mathrm{AMOD}}]$, expected vehicle arrivals $\{n_{i,t'}^{\mathrm{AMOD}}\}_{t'=t+1}^{t+H^{\mathrm{P}}}$ over planning horizon $H^{\mathrm{P}}$, current queue length $q_{i,t}^{\mathrm{AMOD}}$, aggregated demand from the last time step $d_{i,t-1}^{\mathrm{AMOD}}=\sum_{j \in \mathcal{V}} d_{i,j,t-1}^{\mathrm{AMOD}}$, and aggregated price levels from the last time step $p_{i,t-1}^{\mathrm{AMOD}}=\sum_{j \in \mathcal{V}} p_{i,j,t-1}^{\mathrm{AMOD}}$. Regional service availability regulation adds the penalty level $\phi^{\mathrm{SA}}$ and regional service rates measured in the preceding time step:
\begin{equation}
    s_{i,t-1}^{\mathrm{AMOD}}=
    \begin{cases}
        \displaystyle\min\left\{1,\frac{\sum_{j \in \mathcal{V}} x_{i,j,t-1}^{\mathrm{AMOD}}}{\sum_{j \in \mathcal{V}} d_{i,j,t-1}}\right\},
        & \sum_{j \in \mathcal{V}} d_{i,j,t-1}>0,\\
        1, & \text{otherwise},
    \end{cases}
\end{equation}
where $x_{i,j,t-1}^{\mathrm{AMOD}}$ is the passenger flow served by \gls{acr:amod}. Normalizing by total regional traveler demand, rather than only demand choosing \gls{acr:amod}, discourages the operator from meeting the service target solely by raising prices to suppress \gls{acr:amod} demand and instead encourages pricing and rebalancing that increase the number of travelers served. Because served flows may include requests queued in earlier time steps, the service rate is capped. Under surge pricing regulation, the state additionally includes regional price factors $\hat p_{i,t-1}^{\mathrm{AMOD}}$ and the penalty level $\phi^{\mathrm{SP}}$.

\textit{Action ($\mathcal{A}$):}
The operator selects the joint action $\mathbf{a}_t^{\mathrm{AMOD}}=[\hat{\mathbf{p}}_t^{\mathrm{AMOD}},\mathbf{w}_t^{\mathrm{AMOD}}]$, where $\hat{\mathbf{p}}_t^{\mathrm{AMOD}}=[\hat p_{i,t}^{\mathrm{AMOD}}]_{i\in\mathcal{V}}$ and $\mathbf{w}_t^{\mathrm{AMOD}}=[w_{i,t}^{\mathrm{AMOD}}]_{i\in\mathcal{V}}$ collect normalized regional price factors and desired idle-vehicle shares, respectively. These satisfy $\hat p_{i,t}^{\mathrm{AMOD}}\in(0,1]$, $w_{i,t}^{\mathrm{AMOD}}\in[0,1]$, and $\sum_{i\in\mathcal{V}}w_{i,t}^{\mathrm{AMOD}}=1$. The desired idle-vehicle shares constrain a rebalancing optimization, while origin-based price factors determine \gls{acr:od}-level fares:
\begin{equation}
p_{i,j,t}^{\mathrm{AMOD}}
=\beta^{\mathrm{P}}\hat p_{i,t}^{\mathrm{AMOD}}\bar p_{i,j,t}^{\mathrm{AMOD}},
\quad \forall i,j\in\mathcal{V},
\label{eq:price}
\end{equation}
where $\bar p_{i,j,t}^{\mathrm{AMOD}}$ is the historical reference fare and $\beta^{\mathrm{P}}=2$ bounds prices at twice that fare. Origin-based pricing reduces action dimensionality and achieves performance comparable to \gls{acr:od}-based pricing in unregulated control \cite{li2025joint}.

\textit{Reward ($R$):}
The operator maximizes profit:
\begin{align}
    r_t^{\mathrm{AMOD}}
    ={}&\sum_{i,j\in\mathcal{V}} x_{i,j,t}^{\mathrm{AMOD}}
    \left(p_{i,j,t}^{\mathrm{AMOD}}-c_{i,j,t}^{\mathrm{AMOD}}\right)\nonumber\\
    &-\sum_{i,j\in\mathcal{V}} y_{i,j,t}^{\mathrm{AMOD}}c_{i,j,t}^{\mathrm{AMOD}}
    -\Pi_t^r,
    \label{eq:static_reward}
\end{align}
where $y_{i,j,t}^{\mathrm{AMOD}}$ denotes rebalancing flows and $\Pi_t^{r}$ denotes the penalty imposed by the active instrument $r\in\{\mathrm{SA},\mathrm{SP}\}$. Regional service availability ($r=\mathrm{SA}$) penalties depend on regional service deficits $\delta_{i,t}^{\mathrm{SA}}$ and penalty level $\phi^\mathrm{SA}$:
\begin{align}
    \Pi_t^{\mathrm{SA}}&=\sum_{i\in\mathcal{V}}\delta_{i,t}^{\mathrm{SA}}\phi^{\mathrm{SA}},
    \label{eq:static_reward_se}\\
    \delta_{i,t}^{\mathrm{SA}}&=\max\left(s^{\mathrm{AMOD},*}-s_{i,t}^{\mathrm{AMOD}},0\right),
    \label{eq:service_deficit}
\end{align}
where $s^{\mathrm{AMOD},*}\in[0,1]$ is the target service rate. Surge pricing ($r=\mathrm{SP}$) penalties apply to each offered trip whose price exceeds the relative threshold $p^{\mathrm{AMOD},*}\in[1,\beta^{\mathrm{P}}]$ and depend on the penalty level $\phi^\mathrm{SP}$ and excess $\delta_{i,t}^{\mathrm{SP}}$:
\begin{align}
    \Pi_t^{\mathrm{SP}}&=\sum_{i\in\mathcal{V}}d_{i,t}\delta_{i,t}^{\mathrm{SP}}\phi^{\mathrm{SP}},\\
    \delta_{i,t}^{\mathrm{SP}}&=\max\left(\beta^{\mathrm{P}}\hat p_{i,t}^{\mathrm{AMOD}}-p^{\mathrm{AMOD},*},0\right),
    \label{eq:surge_price_excess}
\end{align}
with $d_{i,t}=\sum_{j \in \mathcal{V}} d_{i,j,t}$ denoting trip offers originating in region $i$.

\textit{Transition ($T$):}
Adjacency and penalties remain fixed; prices follow Eq.~\eqref{eq:price}. Demand, queues, and vehicle availability evolve as follows.

\textit{Demand Generation and Mode Choice:} Demand is sampled as $d_{i,j,t}\sim\mathrm{Poisson}(\beta^{\mathrm{D}}\bar d_{i,j,t})$, where historical taxi demand $\bar d_{i,j,t}$ is scaled by $\beta^{\mathrm{D}}=1.2$ to capture latent demand that may switch from \gls{acr:pt}. Generated travelers choose between modes using a binary logit model adapted from \cite{toft2026competitive} with utilities:
\begin{align}
    U_{i,j,t}^{\mathrm{PT}}
    &=-\beta^{\mathrm{T}}\rho\tau_{i,j,t}^{\mathrm{PT}}-p_{i,j,t}^{\mathrm{PT}},\\
    U_{i,j,t}^{\mathrm{AMOD}}
    &=\beta^0-\beta^{\mathrm{T}}\rho\tau_{i,j,t}^{\mathrm{AMOD}}-p_{i,j,t}^{\mathrm{AMOD}}.
\end{align}
Here, travel times are converted to hours, $\rho$ is the mean hourly wage, $\beta^{\mathrm{T}}=0.71$ weights travel time \cite{khulbe2023probabilisticsimulation}, and $\beta^0$ captures baseline \gls{acr:amod} preference.
Bernoulli draws using the logit probabilities yield mode-specific demands $d_{i,j,t}^\mathrm{AMOD}$ and $d_{i,j,t}^\mathrm{PT}$. We calibrate $\beta^0$ so that, at reference fares $\bar{p}_{i,j,t}^{\mathrm{AMOD}}$, an average share of $1/\beta^{\mathrm{D}}$ travelers selects \gls{acr:amod}, reproducing historical demand $\bar{d}_{i,j,t}$ in expectation.

\textit{Queue and Vehicle Dynamics:} For each $i \in \mathcal{V}$, queue lengths $q_{i,t}^{\mathrm{AMOD}}$ and idle-vehicle availability $n_{i,t}^{\mathrm{AMOD}}$ evolve as
 \begin{align}
     q_{i,t+1}^{\mathrm{AMOD}}
     ={}&q_{i,t}^{\mathrm{AMOD}}\nonumber\\
     &+\sum_{j\in\mathcal{V}}
     \left(d_{i,j,t}^{\mathrm{AMOD}}-x_{i,j,t}^{\mathrm{AMOD}}-z_{i,j,t}^{\mathrm{AMOD},\omega}\right),
     \label{eq:amod_queue}\\
     n_{i,t+1}^{\mathrm{AMOD}}
     ={}&n_{i,t}^{\mathrm{AMOD}}
     +\sum_{j\in\mathcal{V}}v_{j,i,t}^{\mathrm{AMOD}}\nonumber\\
     &-\sum_{j\in\mathcal{V}}
     \left(x_{i,j,t}^{\mathrm{AMOD}}+y_{i,j,t}^{\mathrm{AMOD}}\right),
     \label{eq:amod_vehicles}
 \end{align}
where $z_{i,j,t}^{\mathrm{AMOD},\omega}$ counts requests abandoned after exceeding waiting threshold $\omega$ and $v_{j,i,t}^{\mathrm{AMOD}}$ counts vehicles arriving after passenger service or rebalancing.

\subsubsection{Dynamic Regulation as \gls{acr:posg}}

Dynamic regional service availability regulation yields a \gls{acr:posg} with agents $\mathcal{I}=\{\mathrm{AMOD},\mathrm{REG}\}$, shared state space $\mathcal{S}$ and transition $T$, and for each agent $\ell \in \mathcal{I}$ action spaces $\mathcal{A}^{\ell}$, observation spaces $\mathcal{O}^{\ell}$, and rewards $R^{\ell}$ \cite{albrecht2024marlbook, hansen2004posg}.

The regulator ($\ell = \mathrm{REG}$) acts at time steps $\theta_k=k\Delta^{\mathrm{REG}}$, with $k\in\mathcal{K}=\{0,\dots,K-1\}$ and number of regulatory decision points $K=\lceil H/\Delta^{\mathrm{REG}}\rceil$. At each regulatory decision point $k$, it observes information from the observation window $\Omega_k = \{\max(0,\theta_k-\Delta^{\mathrm{REG}}+1), \dots, \theta_k\}$ and sets penalty levels that remain fixed throughout the subsequent regulatory interval $\Theta_k = \{\theta_k, \dots, \min(\theta_k + \Delta^{\mathrm{REG}}-1, H-1)\}$.

\textit{State ($\mathcal{S}$) and Transition ($T$):}
Regional and time-dependent penalties $\phi_{i,t}^{\mathrm{SA}}$ replace $\phi^{\mathrm{SA}}$ in the static case and are updated by Eq.~\eqref{eq:dynamic_penalty_level}. 

\textit{Operator Components ($\mathcal{O}^{\mathrm{AMOD}}$, $\mathcal{A}^{\mathrm{AMOD}}$, $R^{\mathrm{AMOD}}$):}
The operator observes the full state, $\mathbf{o}_t^{\mathrm{AMOD}}=\mathbf{s}_t$, and retains its actions and reward, with $\Pi_t^{\mathrm{SA}}=\sum_i\delta_{i,t}^{\mathrm{SA}}\phi_{i,t}^{\mathrm{SA}}$.

\textit{Regulator Observation ($\mathcal{O}^{\mathrm{REG}}$):}
The regulator receives only a partial observation $\mathbf{o}_k^{\mathrm{REG}}$ of the state at each $k \in \mathcal{K}$, including: adjacency matrix $\mathbf{A}$, regulatory decision point $k$, and for each $i \in \mathcal{V}$: service rates of the current observation window $\{s_{i,t}^{\mathrm{AMOD}}\}_{t\in\Omega_k}$, and previous penalty factors $\hat\phi_{i,k-1}^{\mathrm{SA}}$. 

\textit{Regulator Action ($\mathcal{A}^{\mathrm{REG}}$):}
The action $\mathbf{a}_k^{\mathrm{REG}}=[\hat{\phi}_{i,k}^{\mathrm{SA}}]_{i\in\mathcal{V}}$ sets 
regional service availability penalty factors $\hat{\phi}_{i,k}^{\mathrm{SA}}\in[0,1]$, scaled by upper bound $\beta^{\mathrm{SA}}$, to obtain the penalty level applied throughout the regulatory interval $\Theta_k$:
\begin{equation}
\phi_{i,t}^{\mathrm{SA}}=\beta^{\mathrm{SA}}\hat{\phi}_{i,k}^{\mathrm{SA}},
\qquad i\in\mathcal{V},\ t\in\Theta_k.
\label{eq:dynamic_penalty_level}
\end{equation}

\textit{Regulator Reward ($R^{\mathrm{REG}}$):}
The regulator's reward over $\Theta_k$ balances normalized service deficits and penalty levels:
\begin{equation}
    r_k^{\mathrm{REG}}=-\sum_{t\in\Theta_k}
    \left(
    \sum_{i\in\mathcal{V}}\frac{\delta_{i,t}^{\mathrm{SA}}}{s^{\mathrm{AMOD},*}}
    +\lambda\sum_{i\in\mathcal{V}}\hat{\phi}_{i,k}^{\mathrm{SA}}
    \right),
    \label{eq:dynamic_reward_reg}
\end{equation}
with weight $\lambda\in[0,1]$.
The reward reflects the regulator's aim of inducing compliance with the smallest possible penalties. The first term penalizes service deficits, while the second penalizes penalty levels, with $\lambda$ steering the trade-off. We penalize levels rather than realized payments because payments vanish under compliance regardless of the penalty level and thus provide no incentive for the regulator to identify minimal effective penalties.
\subsection{Learning Setup and Model Architecture} \label{sec:learning_setup}

The operator is trained using \gls{acr:sac} due to the continuous action space and off-policy sample efficiency \cite{haarnoja2018sac}. Under dynamic regulation, the operator and regulator learn independently, introducing non-stationarity \cite{albrecht2024marlbook}. We therefore train the regulator with on-policy \gls{acr:a2c} \cite{mnih2016a2c}. To stabilize joint training, the operator is pretrained with static penalties varied between episodes. 

Following \cite{gammelli2021gnnrl,li2025joint}, actor and critic networks use one \gls{acr:gcn} layer and three \gls{acr:fc} layers, with the two hidden layers of size $h$. Operator inputs are $\boldsymbol{\xi}_t^{\mathrm{AMOD}}=\mathbf{s}_t$ under static and $\boldsymbol{\xi}_t^{\mathrm{AMOD}}=\mathbf{o}_t^{\mathrm{AMOD}}$ under dynamic regulation. Regulator inputs are $\mathbf{o}_k^{\mathrm{REG}}$. Critics sum regional embeddings to estimate $Q_{\psi}^{\mathrm{AMOD}}(\boldsymbol{\xi}_t^{\mathrm{AMOD}},\mathbf{a}_t^{\mathrm{AMOD}})$ and $V_{\psi}^{\mathrm{REG}}(\mathbf{o}_k^{\mathrm{REG}})$. Actors define stochastic policies $\pi_{\vartheta}^{\mathrm{AMOD}}(\cdot \mid \boldsymbol{\xi}_t^{\mathrm{AMOD}})$ and $\pi_{\vartheta}^{\mathrm{REG}}(\cdot \mid \mathbf{o}_k^{\mathrm{REG}})$, using Softplus outputs for positive distribution parameters. The operator produces two parameters per region to sample price factors $\hat p_{i,t}^{\mathrm{AMOD}}\sim\mathrm{Beta}(\alpha_{i,t}^{p},\beta_{i,t}^{p})$ and one concentration parameter per region, collected in $\boldsymbol{\gamma}_t$, to sample idle-vehicle shares $\mathbf{w}_t^{\mathrm{AMOD}}\sim\mathrm{Dirichlet}(\boldsymbol{\gamma}_t)$. The regulator produces two parameters per region to sample penalty factors $\hat\phi_{i,k}^{\mathrm{SA}}\sim\mathrm{Beta}(\alpha_{i,k}^{\phi},\beta_{i,k}^{\phi})$. The model architecture and training hyperparameters are reported in Secs.~S-I and~S-II of the supplementary material.


\section{Experiments} \label{chap:results}

\begin{figure}[!t]
    \centering
    \includegraphics[width=0.35\linewidth]{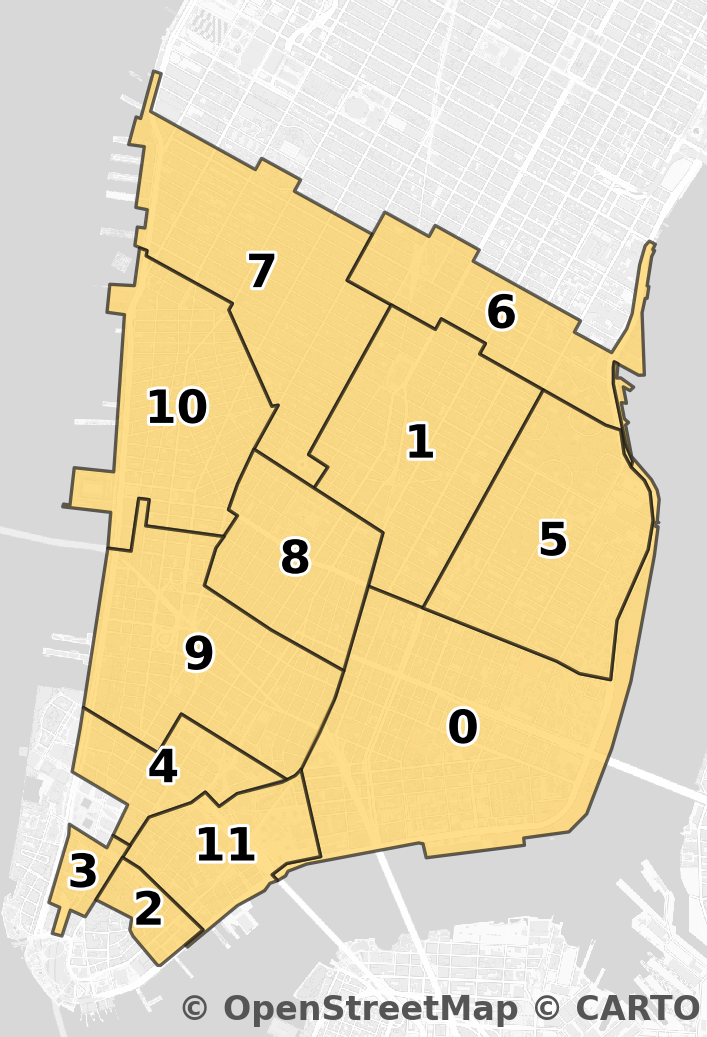}
    \caption[Simulation area]{Simulation area in southern Manhattan, discretized into 12 regions.}
    \label{fig:simulation_area}
\end{figure}

In this section, we evaluate static surge pricing regulation and static and dynamic regional service availability regulation, examining their effectiveness, operator adaptation, and trade-offs for the regulator, operator, and travelers.\footnote{See \url{https://github.com/korbiniankuegler/rl-regulating-amod} for more material and code.} The empirical analysis is based on a scenario implemented in an adapted version of the simulator proposed by \cite{gammelli2022gmrlfortransferable}. The scenario represents 12 ZIP code regions in southern Manhattan (Fig.~\ref{fig:simulation_area}) on 20 September 2013, from 15:00 to 20:59, over 120 three-minute time steps. Reference demand, \gls{acr:amod} fares, and \gls{acr:amod} travel times are derived from historical NYC taxi records \cite{illinoisdatabank}, excluding passenger and rebalancing trips within the same region. Corresponding \gls{acr:pt} travel times are obtained using the FleetPy router \cite{Engelhardt.2026} and NYC bus and subway GTFS data \cite{mta_bus_gtfs_2013,mta_subway_gtfs_2013}. More details on the scenario construction can be found in Sec.~S-III of the supplementary material. 

The operator manages 850 vehicles at an operating cost of US\$18 per vehicle-hour \cite{tresca2025robotaxicoordination}. Travelers have an hourly wage of US\$22.77 \cite{toft2026competitive} and a maximum waiting time of six minutes. The scenario spans afternoon off-peak conditions (15:00--18:59) and evening peak demand (19:00--20:59). To characterize the scenario, we run 100 simulation samples with \gls{acr:amod} prices fixed at historical reference fares. Across both periods, southern regions 2, 3, 4, and 11 have the lowest demand and comparatively low \gls{acr:amod} mode shares. Detailed scenario characteristics are provided in Sec.~S-IV of the supplementary material.

Models are trained for 20,000 episodes, by which point the training rewards have stabilized, and evaluated over ten independent test runs. We report means with standard deviations in parentheses. Reported price factors include the scaling by $\beta^{\mathrm{P}}$, i.e., $\beta^{\mathrm{P}}\hat p_{i,t}^{\mathrm{AMOD}}\in(0,2]$. 

\subsection{Surge Pricing Regulation} \label{sec:results_surge_pricing}
We penalize offered prices above 120\% of reference fares ($p^{\mathrm{AMOD},*}=1.2$) and compare penalty levels $\phi^{\mathrm{SP}}\in\{0,1,5,10\}$, measured in US\$ per unit of price factor excess. Zero denotes no regulation.

\subsubsection{Aggregate Performance Analysis} \label{sec:results_surge_pricing_performance}
Despite a mean price factor of 1.09, the unregulated operator offers surge prices to 22\% of travelers, averaging 127\% of reference fares (Table~\ref{tab:surge_static_results}). Increasing penalties reduces this share to 17\%, 6\%, and 1\%, while the average price factor excess conditional on being affected falls from 0.07 to 0.01. Static regulation thus suffices to nearly eliminate excessive prices while generating negligible penalty payments.


\begin{table}[!ht]
\centering
\caption{Performance metrics under static surge pricing regulation.}
\label{tab:surge_static_results}
\footnotesize
\setlength{\tabcolsep}{2pt}

\begin{tabular}{@{}lrrrr@{}}
\toprule
& \multicolumn{4}{c}{\textbf{Penalty level $\phi^\mathrm{SP}$} (US\$)} \\
\cmidrule(lr){2-5}
\textbf{Metric} & \textbf{0} & \textbf{1} & \textbf{5} & \textbf{10} \\
\midrule

\multicolumn{5}{@{}l}{\textbf{Regulator}} \\

Affected travelers (\#)
& 4,965 (144) & 3,750 (182) & 1,279 (187) & 296 (113) \\

Affected share (\%)
& 22 (1) & 17 (1) & 6 (1) & 1 (1) \\

Avg. price factor excess
& 0.07 (0.00) & 0.06 (0.00) & 0.03 (0.00) & 0.01 (0.00) \\

Penalties (US\$)
& 0 (0) & 230 (11) & 170 (27) & 28 (10) \\

\midrule
\multicolumn{5}{@{}l}{\textbf{Operator}} \\

Reward (k US\$)
& 129.5 (0.7) & 128.7 (0.7) & 127.0 (0.7) & 126.5 (0.7) \\

Rebalancing trips (\#)
& 1,896 (64) & 1,928 (63) & 1,946 (55) & 1,961 (75) \\

Avg. price factor
& 1.09 (0.00) & 1.06 (0.00) & 1.02 (0.00) & 1.02 (0.00) \\

Served demand (\#)
& 16,157 (67) & 16,374 (70) & 16,652 (64) & 16,609 (75) \\

Lost demand (\#)
& 309 (39) & 379 (37) & 601 (46) & 667 (52) \\

Avg. wait (min)
& 0.51 (0.04) & 0.55 (0.04) & 0.62 (0.04) & 0.63 (0.04) \\

Vehicle utilization (\%)
& 63 (0) & 64 (0) & 66 (0) & 65 (0) \\

\midrule
\multicolumn{5}{@{}l}{\textbf{Traveler}} \\

Realized utility (k US\$)
& -116.6 (1.0) & -114.1 (0.8) & -110.1 (1.0) & -109.9 (0.9) \\

AMoD mode share (\%)
& 73 (0) & 75 (0) & 77 (0) & 77 (0) \\

\bottomrule
\end{tabular}

\vspace{0.4em}
\begin{minipage}{\columnwidth}
\footnotesize
\end{minipage}
\end{table}

The operator responds primarily by lowering prices. Rebalancing increases slightly, suggesting partial compensation for the reduced ability to steer demand through prices. In the representative run shown in Table~\ref{tab:surge_utilization_hourly}, peak-hour vehicle utilization already reaches 97\% without regulation, limiting further vehicle repositioning. With lower prices, served demand rises. The additional demand does not offset lower fares and operator reward decreases by up to 2.3\%. Moreover, lost demand due to travelers exceeding their waiting threshold more than doubles, suggesting that suppressing surge prices pushes the operator closer to its capacity limits.

\begin{table}[!ht]
\centering
\caption[Hourly vehicle utilization under static surge pricing regulation]{Vehicle utilization by hour in percent under static surge pricing regulation.}
\label{tab:surge_utilization_hourly}
\footnotesize
\begin{tabular}{lrrrrrr}
\toprule
& \multicolumn{6}{c}{\textbf{Hour of day}} \\
\cmidrule(lr){2-7}
\textbf{Penalty level}
& \textbf{15}
& \textbf{16}
& \textbf{17}
& \textbf{18}
& \textbf{19}
& \textbf{20} \\
\midrule
$\phi^\mathrm{SP}=0$  
& 36 & 27 & 40 & 82 & 97 & 97 \\
$\phi^\mathrm{SP}=1$  
& 37 & 28 & 41 & 84 & 97 & 98 \\
$\phi^\mathrm{SP}=5$  
& 40 & 30 & 44 & 86 & 97 & 98 \\
$\phi^\mathrm{SP}=10$ 
& 40 & 30 & 43 & 85 & 97 & 98 \\
\bottomrule
\end{tabular}

\vspace{0.5em}
\begin{minipage}{\linewidth}
\footnotesize
\centering
\textit{Notes:} Values are shown for one representative evaluation run.
\end{minipage}
\end{table}

Nevertheless, total realized traveler utility, accounting for waiting and fallback to \gls{acr:pt}, improves by up to US\$6,700, as lower fares outweigh service-quality losses on average. Static surge pricing regulation therefore benefits travelers overall but increases financial and operational pressure on the operator and shifts demand from \gls{acr:pt} to \gls{acr:amod}. By choosing the penalty level, the regulator can prioritize these effects.

\textit{Operator sensitivity to regulation:}
At $\phi^{\mathrm{SP}}\in\{1,5\}$, an unregulated operator policy evaluated under penalties earns slightly higher mean reward than the corresponding operator trained under regulation (Table~\ref{tab:surge_reaction}). Adaptation improves mean reward only at $\phi^{\mathrm{SP}}=10$. The over-adjustment at lower penalty levels is likely caused by high penalties during early exploration, encouraging overly conservative pricing (see Sec.~S-V of the supplementary material). Observed responses may therefore partly reflect training artifacts. Warm-starting from an unregulated policy could be investigated to address this limitation.

\begin{table}[!ht]
\centering
\caption[Operator sensitivity to static surge pricing regulation]{Operator reward comparison by evaluation and training penalty level under static surge pricing.}
\label{tab:surge_reaction}
\footnotesize
\begin{tabular}{llr}
\toprule
\textbf{Evaluation penalty} & \textbf{Training penalty} & \textbf{Reward} (k US\$) \\
\midrule
$\phi^\mathrm{SP}=0$  
& $\phi^\mathrm{SP}=0$  
& \textbf{129.5 (0.7)} \\
\midrule

$\phi^\mathrm{SP}=1$  
& $\phi^\mathrm{SP}=1$  
& 128.7 (0.7) \\
& $\phi^\mathrm{SP}=0$  
& \textbf{129.1 (0.7)} \\
\midrule

$\phi^\mathrm{SP}=5$  
& $\phi^\mathrm{SP}=5$  
& 127.0 (0.7) \\
& $\phi^\mathrm{SP}=0$  
& \textbf{127.6 (0.7)} \\
\midrule

$\phi^\mathrm{SP}=10$ 
& $\phi^\mathrm{SP}=10$ 
& \textbf{126.5 (0.7)} \\
& $\phi^\mathrm{SP}=0$  
& 125.9 (0.8) \\
\bottomrule
\end{tabular}

\vspace{0.5em}
\begin{minipage}{\linewidth}
\footnotesize
\centering
\textit{Notes:} Best-performing policy for each evaluation penalty level is in bold.
\end{minipage}
\end{table}

\subsubsection{Regional Performance Analysis} \label{sec:results_surge_pricing_policy}
To explain the aggregate effects, we compare operator policies, regulation outcomes, and traveler outcomes at regional and hourly levels, using representative evaluation runs without regulation and with $\phi^{\mathrm{SP}}=10$.

\begin{figure*}[!t]
    \centering
    \subfloat[Affected traveler share (\%), unregulated\label{fig:surge_affected_unreg}]{%
        \includegraphics[width=0.32\linewidth]{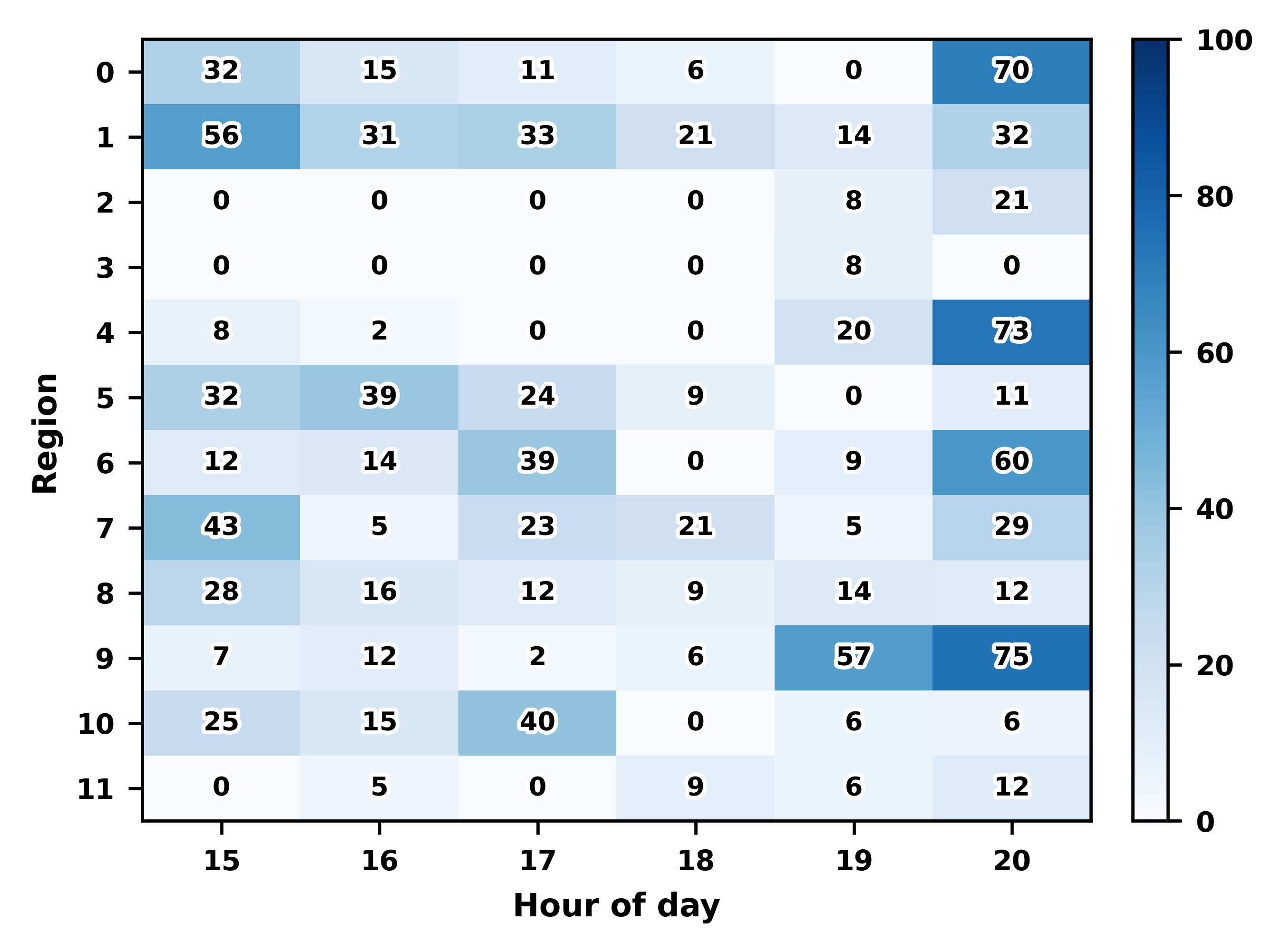}%
    }\hfill
    \subfloat[Mean price factor, unregulated\label{fig:surge_price_unreg}]{%
        \includegraphics[width=0.32\linewidth]{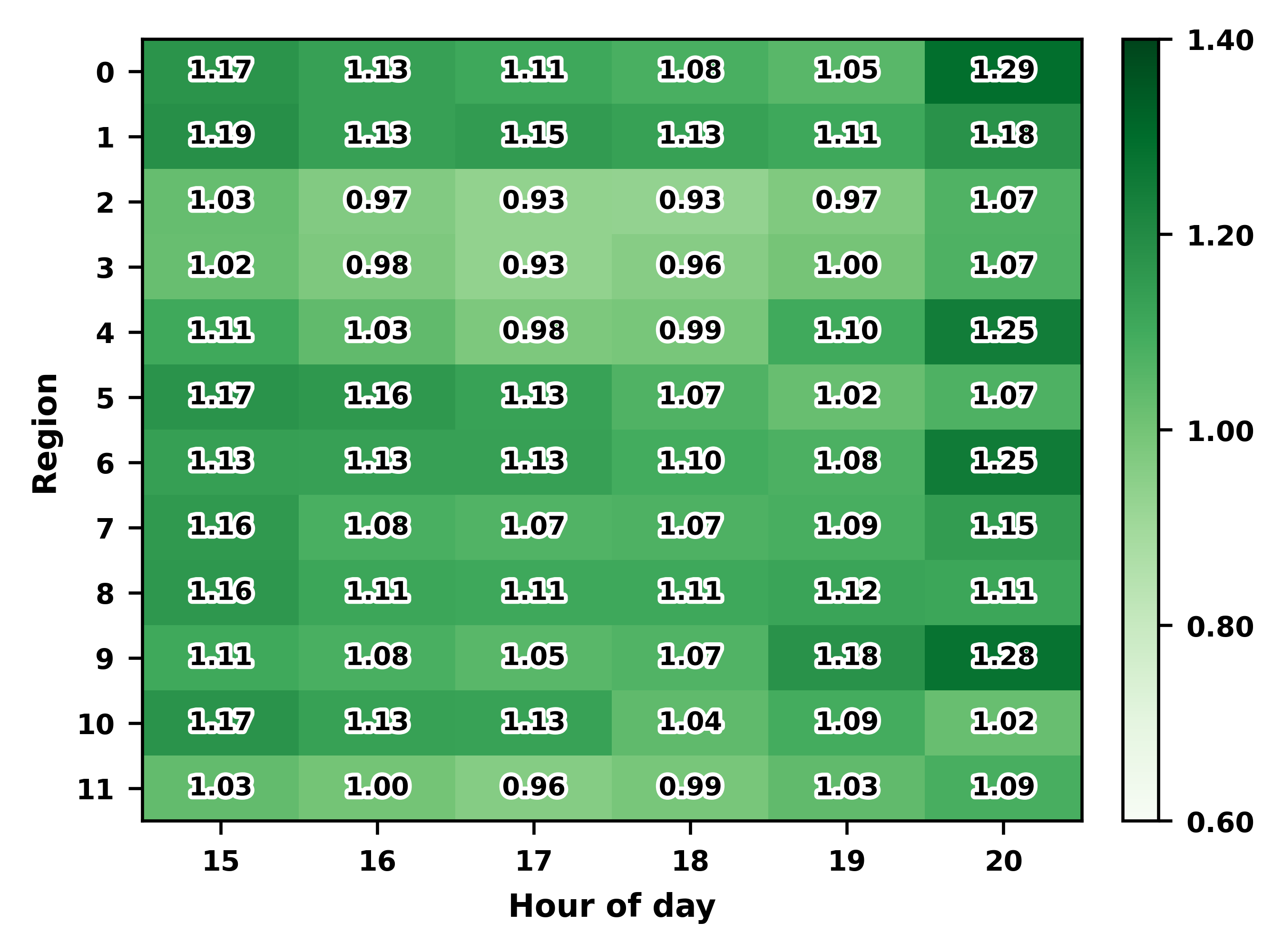}%
    }\hfill
    \subfloat[Realized utility per traveler (US\$), unregulated\label{fig:surge_utility_unreg}]{%
        \includegraphics[width=0.32\linewidth]{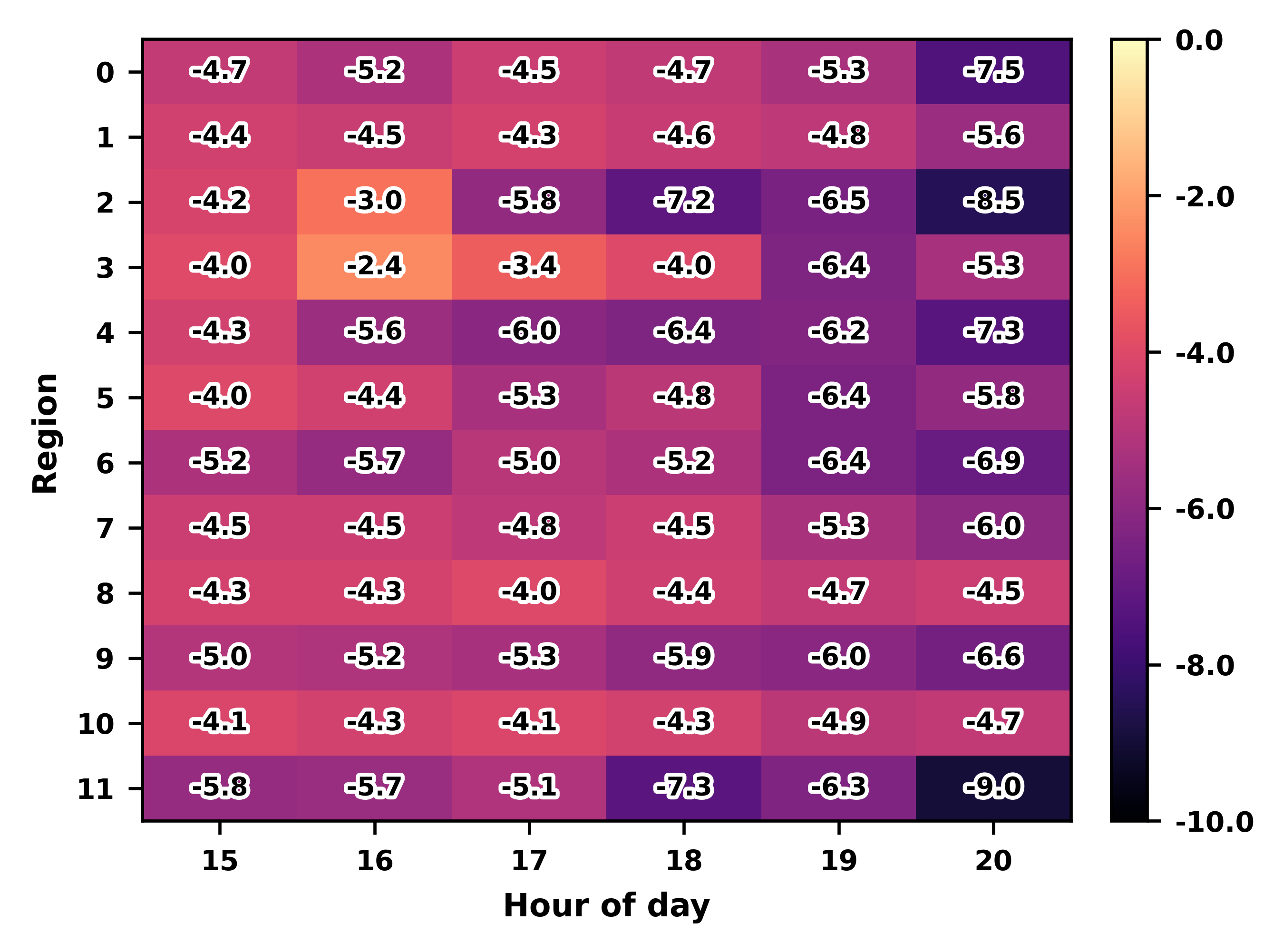}%
    }

    \vspace{-1em}

    \subfloat[Affected traveler share (\%), regulated\label{fig:surge_affected_reg}]{%
        \includegraphics[width=0.32\linewidth]{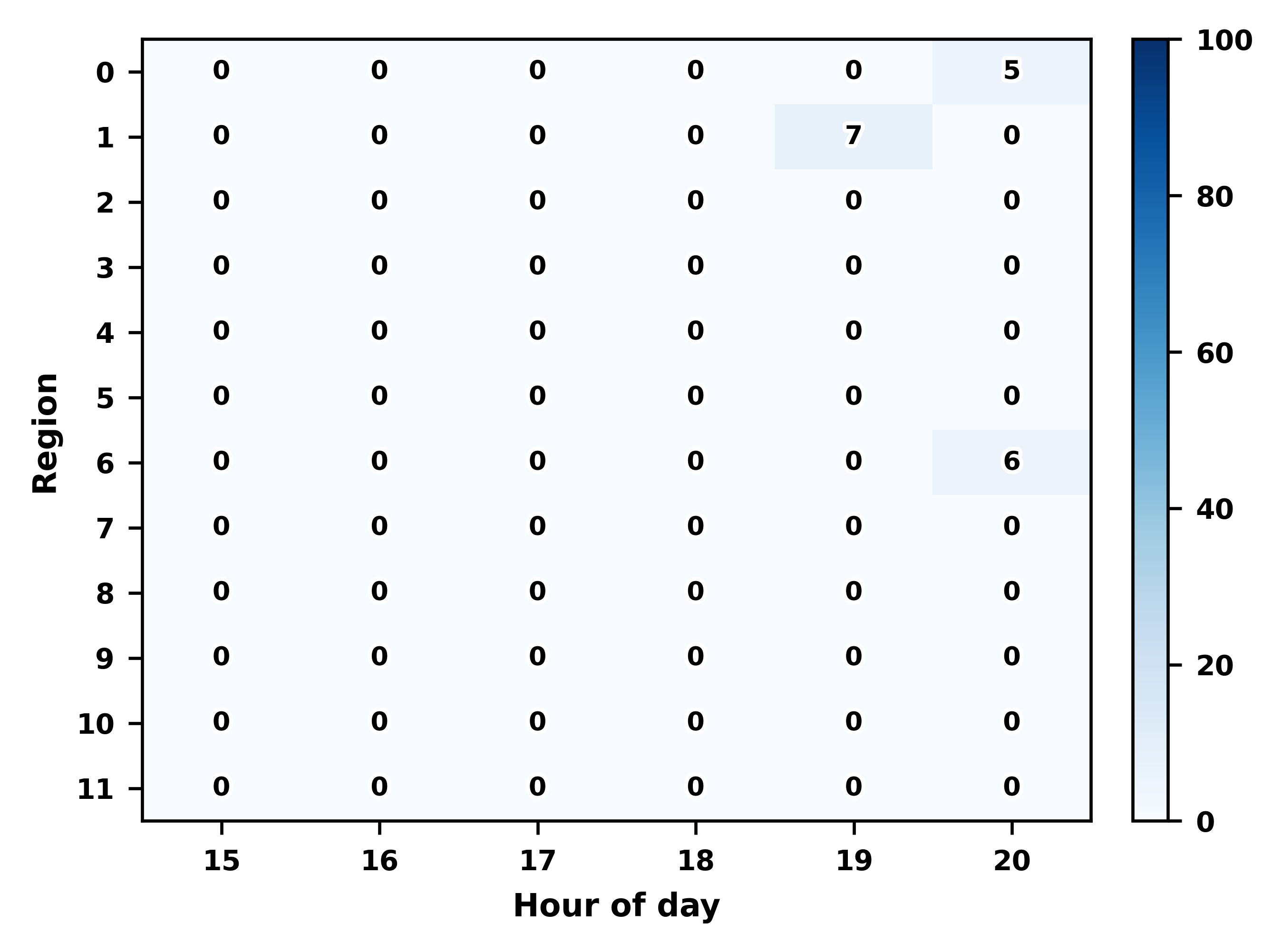}%
    }\hfill
    \subfloat[Mean price factor, regulated\label{fig:surge_price_reg}]{%
        \includegraphics[width=0.32\linewidth]{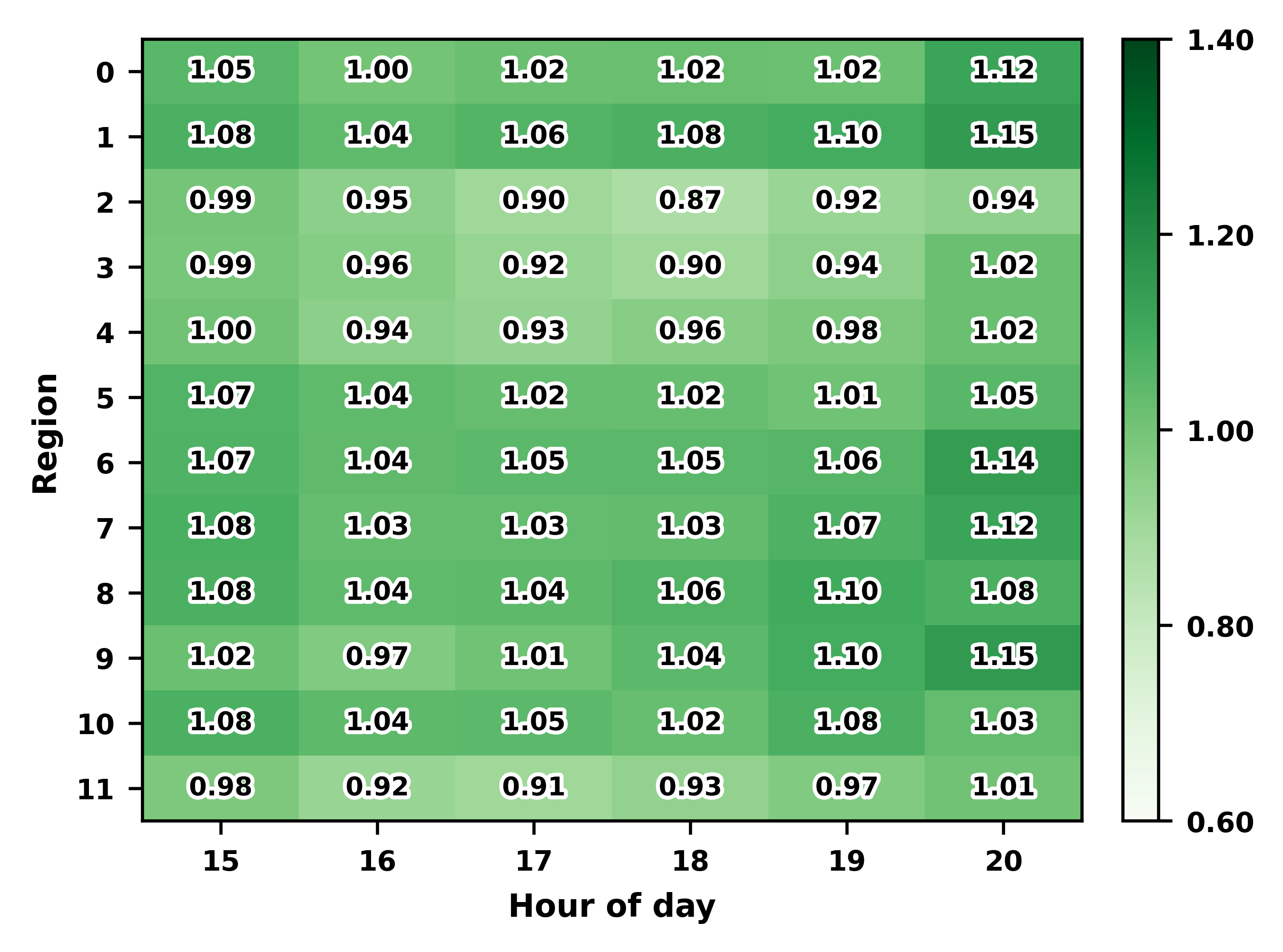}%
    }\hfill
    \subfloat[Realized utility per traveler (US\$), regulated\label{fig:surge_utility_reg}]{%
        \includegraphics[width=0.32\linewidth]{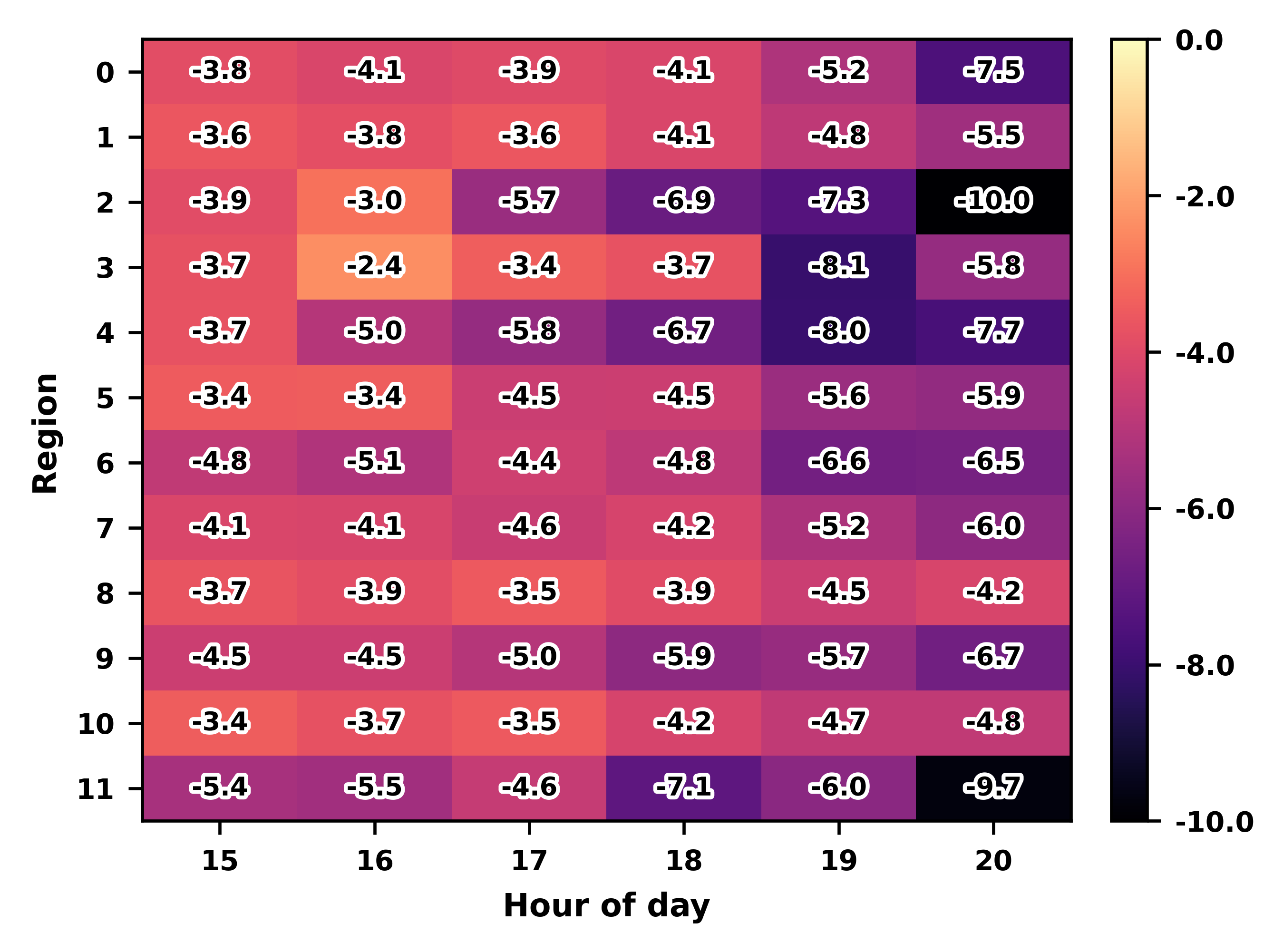}%
    }

    \caption[Pricing policy, traveler outcomes, and regulation outcomes under static surge pricing regulation]{Regulation outcomes, pricing policy, and traveler outcomes with and without static surge pricing regulation by region and hour in one representative evaluation run. Travelers count as affected if the surge pricing factor is above the threshold.}
    \label{fig:surge_policy_response}
\end{figure*}

Without regulation, surge pricing is strongest during the evening peak, affecting 60--75\% of travelers in regions 0, 4, 6, and 9 in the final hour (Fig.~\ref{fig:surge_policy_response}). Southern regions 2, 3, and 11 remain less affected, consistent with their higher price sensitivity. Regions 5, 8, and 10, which exhibit net demand inflows during peak hours, also face less surge pricing. This suggests that the operator may use lower prices to encourage trips from these regions toward regions needing vehicles, reducing the need for costly rebalancing.

As shown in the second row of Fig.~\ref{fig:surge_policy_response}, regulation lowers prices while largely preserving the spatial pricing pattern. Prices also decline where surge pricing was absent, consistent with the over-adjustment discussed above. Rebalancing changes remain limited, particularly during the evening peak with high vehicle utilization.

Off-peak travelers generally benefit from lower fares without substantial service-quality losses (Fig.~\ref{fig:surge_policy_response}). However, during peak hours, waiting times and lost demand increase, particularly where surge prices previously rationed demand (see Sect.~S-VI of the supplementary material). Realized utility consequently worsens in several region-hour combinations, often where baseline utility was already low. These patterns suggest that surge pricing mainly captures traveler surplus off-peak but also rations scarce capacity during peaks. Suppressing it can shift scarcity from monetary costs to service-quality losses and intensify disparities despite improving average traveler utility.

\subsection{Regional Service Availability Regulation} \label{sec:results_spatial_equity}
Regional service availability regulation penalizes service rates below $s^{\mathrm{AMOD},*}=0.6$. The regulator adjusts penalty levels hourly ($\Delta^{\mathrm{REG}}=20$), with upper bound $\beta^{\mathrm{SA}}=200$ US\$ per unit of service deficit. We evaluate $\lambda\in\{1,0.1,0.05,0.01\}$, where higher weights penalize intervention more strongly (Eq.~\eqref{eq:dynamic_reward_reg}).

\subsubsection{Static and Dynamic Regulation Comparison} \label{sec:results_spatial_equity_comparison}
We compare dynamic policies with static benchmarks matching their mean penalty level across regions and time, $\bar{\phi}^{\mathrm{SA}}=(\left|\mathcal{V}\right|H)^{-1}\sum_{i\in\mathcal{V}}\sum_{t=0}^{H-1}\phi_{i,t}^{\mathrm{SA}}$, alongside extremes $\phi^{\mathrm{SA}}\in\{0,200\}$ (Table~\ref{tab:spatial_comparison}). Dynamic regulation achieves at least the same mean regulator reward as the matched benchmarks, with improvements of 15.6\% at $\lambda=0.1$ and 10.5\% at $\lambda=0.05$. These gains suggest value in adjusting penalties across regions and time beyond selecting their average intensity.

\begin{table}[!ht]
\centering
\caption[Static and dynamic spatial equity regulation performance comparison]{Static and dynamic spatial equity regulation performance comparison.}

\label{tab:spatial_comparison}
\footnotesize
\setlength{\tabcolsep}{2.5pt}
\begin{tabular}{@{}llcrrr@{}}
\toprule
\textbf{$\lambda$} 
& \textbf{Regulation} 
& \makecell[t]{\textbf{Penalty}\\\textbf{level}}
& \makecell[t]{\textbf{Operator}\\\textbf{reward}\\(US\$)}
& \makecell[t]{\textbf{Traveler}\\\textbf{utility}\\(k US\$)}
& \makecell[t]{\textbf{Regulator}\\\textbf{reward}\\(k US\$)} \\
\midrule

\multirow[t]{4}{*}{1.0}
& Dynamic & $\bar{\phi}^{\mathrm{SA}}=4$ 
& 125.6 (0.8) & -112.0 (0.9) & -184 (7) \\

& Static & $\phi^{\mathrm{SA}}=0$ 
& 129.5 (0.9) & -115.1 (0.8) & \textbf{-164 (7)} \\

& Static & $\phi^{\mathrm{SA}}=4$ 
& 128.8 (0.7) & -114.0 (1.0) & -184 (8) \\

& Static & $\phi^{\mathrm{SA}}=200$ 
& 118.3 (0.8) & -110.1 (0.9) & -1,521 (5) \\

\midrule

\multirow[t]{4}{*}{0.1}
& Dynamic & $\bar{\phi}^{\mathrm{SA}}=22$ 
& 124.0 (0.7) & -109.5 (0.9) & \textbf{-124 (5)} \\

& Static & $\phi^{\mathrm{SA}}=0$ 
& 129.5 (0.9) & -115.1 (0.8) & -164 (7) \\

& Static & $\phi^{\mathrm{SA}}=22$ 
& 127.5 (0.6) & -114.3 (0.9) & -147 (6) \\

& Static & $\phi^{\mathrm{SA}}=200$ 
& 118.3 (0.8) & -110.1 (0.9) & -225 (5) \\

\midrule

\multirow[t]{4}{*}{0.05}
& Dynamic & $\bar{\phi}^{\mathrm{SA}}=47$ 
& 125.1 (0.7) & -109.2 (0.9) & \textbf{-119 (5)} \\

& Static & $\phi^{\mathrm{SA}}=0$ 
& 129.5 (0.9) & -115.1 (0.8) & -164 (7) \\

& Static & $\phi^{\mathrm{SA}}=47$ 
& 126.1 (0.7) & -114.5 (0.9) & -133 (6) \\

& Static & $\phi^{\mathrm{SA}}=200$ 
& 118.3 (0.8) & -110.1 (0.9) & -153 (5) \\

\midrule

\multirow[t]{4}{*}{0.01}
& Dynamic & $\bar{\phi}^{\mathrm{SA}}=177$ 
& 117.4 (0.7) & -107.8 (0.9) & -96 (4) \\

& Static & $\phi^{\mathrm{SA}}=0$ 
& 129.5 (0.9) & -115.1 (0.8) & -164 (7) \\

& Static & $\phi^{\mathrm{SA}}=177$ 
& 118.6 (0.7) & -109.1 (1.0) & -98 (4) \\

& Static & $\phi^{\mathrm{SA}}=200$ 
& 118.3 (0.8) & -110.1 (0.9) & \textbf{-95 (5)} \\

\bottomrule
\end{tabular}

\vspace{0.5em}
\begin{minipage}{\linewidth}
\footnotesize
\centering
\textit{Notes:} Best-performing policy for each lambda is in bold.
\end{minipage}
\end{table}

However, dynamic regulation does not dominate all static settings. No regulation performs best at $\lambda=1$, while maximum static penalties slightly outperform the dynamic policy at $\lambda=0.01$. This likely reflects the regulator’s continuous Beta-distributed policy, which can approximate but does not sample exact boundary actions. Dynamic regulation yields higher traveler utility than all compared static settings for three of four weights, whereas matched static policies yield higher operator rewards. Beyond more binding dynamic incentives, differences in operator rewards may reflect differing learning environments and model selection criteria. Under dynamic regulation, limited operator adjustment may trigger higher future penalties, potentially encouraging more conservative behavior than under fixed penalties, even at similar average penalty levels. Furthermore, under static regulation, we select the operator policy that achieves the highest operator reward during training. Under dynamic regulation, we instead select the operator–regulator policy pair that achieves the highest regulator reward, which may partly explain the operator's lower reward under dynamic regulation.

\subsubsection{Aggregate Performance Analysis} \label{sec:results_spatial_equity_performance}

Without regulation, the mean service rate is 73\%, yet 26\% of regions fall below the 60\% target and the mean service deficit is 7\% (Table~\ref{tab:spatial_dynamic_results}). Under dynamic regulation, decreasing $\lambda$ raises mean penalty levels, reducing service deficits from 6\% to 3\% and affected regions from 24\% to 16\%. The strongest dynamic setting matches maximum static regulation's aggregate availability outcomes but does not eliminate deficits.


\begin{table}[!ht]
\centering
\caption{Performance metrics under spatial equity regulation.}
\label{tab:spatial_dynamic_results}
\footnotesize
\setlength{\tabcolsep}{2pt}

\begin{tabular}{@{}lrrrrrr@{}}
\toprule
& \textbf{Static}
& \multicolumn{4}{c}{\textbf{Dynamic}}
& \textbf{Static} \\
\cmidrule(lr){2-2}
\cmidrule(lr){3-6}
\cmidrule(lr){7-7}
& $\phi^{\mathrm{SA}}$
& \multicolumn{4}{c}{$\lambda$}
& $\phi^{\mathrm{SA}}$\\
\textbf{Metric}
& \multicolumn{1}{c}{\textbf{0}}
& \multicolumn{1}{c}{\textbf{1.0}}
& \multicolumn{1}{c}{\textbf{0.1}}
& \multicolumn{1}{c}{\textbf{0.05}}
& \multicolumn{1}{c}{\textbf{0.01}}
& \multicolumn{1}{c}{\textbf{200}} \\
\midrule

\multicolumn{7}{@{}l}{\textbf{Regulator}} \\

Avg. penalty level (US\$)
& 0 & 4 & 22 & 47 & 177 & 200 \\
& (0.00) & (0.05) & (0.34) & (0.35) & (0.29) & (0.00) \\

Avg. service rate (\%)
& 73 & 74 & 78 & 79 & 81 & 81 \\
& (0) & (1) & (0) & (0) & (0) & (0) \\

Avg. service deficit (\%)
& 7 & 6 & 5 & 4 & 3 & 3 \\
& (0) & (0) & (0) & (0) & (0) & (0) \\

Affected regions (\%)
& 26 & 24 & 19 & 19 & 16 & 16 \\
& (1) & (1) & (1) & (1) & (1) & (1) \\

Penalties (US\$)
& 0 & 371 & 1,114 & 2,065 & 9,230 & 9,710 \\
& (0) & (23) & (79) & (144) & (480) & (608) \\

\midrule
\multicolumn{7}{@{}l}{\textbf{Operator}} \\

Reward (k US\$)
& 129.5 & 125.6 & 124.0 & 125.1 & 117.4 & 118.3 \\
& (0.9) & (0.8) & (0.7) & (0.7) & (0.7) & (0.8) \\

Rebalancing trips (\#)
& 1,923 & 2,065 & 2,128 & 2,046 & 2,062 & 2,078 \\
& (68) & (44) & (61) & (55) & (67) & (69) \\

Avg. price factor
& 1.08 & 1.02 & 0.98 & 0.97 & 0.96 & 0.98 \\
& (0.00) & (0.00) & (0.00) & (0.00) & (0.00) & (0.00) \\

Served demand (\#)
& 16,297 & 16,402 & 16,584 & 16,815 & 16,908 & 16,798 \\
& (83) & (70) & (69) & (66) & (71) & (79) \\

Lost demand (\#)
& 320 & 500 & 418 & 440 & 369 & 315 \\
& (36) & (36) & (45) & (31) & (38) & (36) \\

Avg. wait (min)
& 0.52 & 0.60 & 0.51 & 0.57 & 0.52 & 0.49 \\
& (0.04) & (0.04) & (0.03) & (0.05) & (0.03) & (0.04) \\

Vehicle utilization (\%)
& 64 & 65 & 67 & 68 & 68 & 68 \\
& (0) & (0) & (0) & (0) & (0) & (0) \\

\midrule
\multicolumn{7}{@{}l}{\textbf{Traveler}} \\

Realized utility (k US\$)
& -115.1 & -112.0 & -109.5 & -109.2 & -107.8 & -110.1 \\
& (0.8) & (0.9) & (0.9) & (0.9) & (0.9) & (0.9) \\

AMoD mode share (\%)
& 74 & 76 & 76 & 77 & 77 & 76 \\
& (0) & (0) & (0) & (0) & (0) & (0) \\

\bottomrule
\end{tabular}

\vspace{0.4em}
\begin{minipage}{\linewidth}
\footnotesize
\textit{Notes:} The average penalty level denotes $\bar{\phi}^{\mathrm{SA}}$
for dynamic and $\phi^{\mathrm{SA}}$ for static
regulation. Regulator rewards are not comparable across
$\lambda$ settings.
\end{minipage}
\end{table}

Relative to no regulation, the strongest dynamic setting lowers the mean price factor, increases served demand, and raises vehicle utilization. Rebalancing exceeds the unregulated level but peaks at $\lambda=0.1$, potentially constrained by higher utilization. Lost demand initially rises, then approaches the unregulated level as penalties strengthen despite higher demand.

Lower fares improve total realized traveler utility by up to US\$7,300 but, together with penalty payments, reduce operator reward by up to US\$12,100. Penalty payments reach 8\% of profits at $\lambda=0.01$, yet deficits persist, indicating possible structural constraints or remaining profitability trade-offs. 
Selecting $\lambda$ reflects a normative trade-off between regional service availability and traveler benefits on the one hand, and regulatory intervention intensity and operator profitability on the other. This trade-off becomes increasingly pronounced at high penalty levels, as marginal availability gains diminish while financial pressure on the operator increases. 

\subsubsection{Regional Performance Analysis} \label{sec:results_spatial_equity_policy}
\begin{figure}[!b]
    \centering
    \includegraphics[width=0.32\textwidth]
        {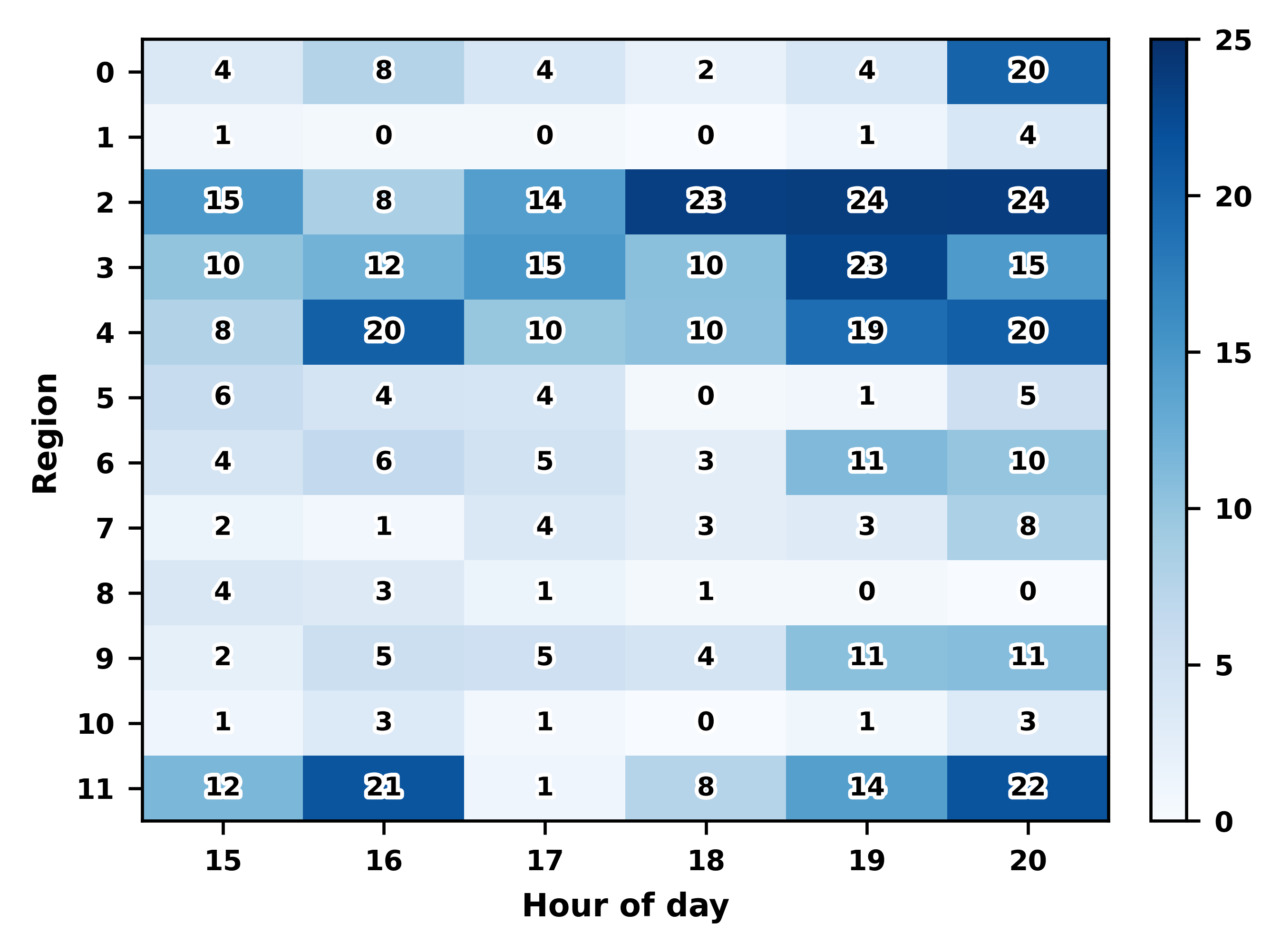}
    \caption{Mean service deficit (\%) without regional service availability regulation.}
    \label{fig:spatial_deficit_unreg}
\end{figure}

Representative runs without regulation and with dynamic regulation at $\lambda=0.05$ show that unregulated deficits concentrate in southern regions 2, 3, 4, and 11 and spread during the evening peak (Fig.~\ref{fig:spatial_deficit_unreg}). These regions combine low demand, low \gls{acr:amod} mode shares at historical reference fares, and high price sensitivity. In Sec.~S-IV of the supplementary material, we show that this pattern is not solely explained by structurally superior \gls{acr:pt} alternatives, but also by pricing conditions the operator can influence. Regions 4 and 11 also exhibit net demand outflows, making service there comparatively unattractive to the operator.

\begin{figure*}[!t]
    \centering
    \subfloat[Mean service deficit (\%)\label{fig:spatial_deficit_reg}]{%
        \includegraphics[width=0.32\linewidth]{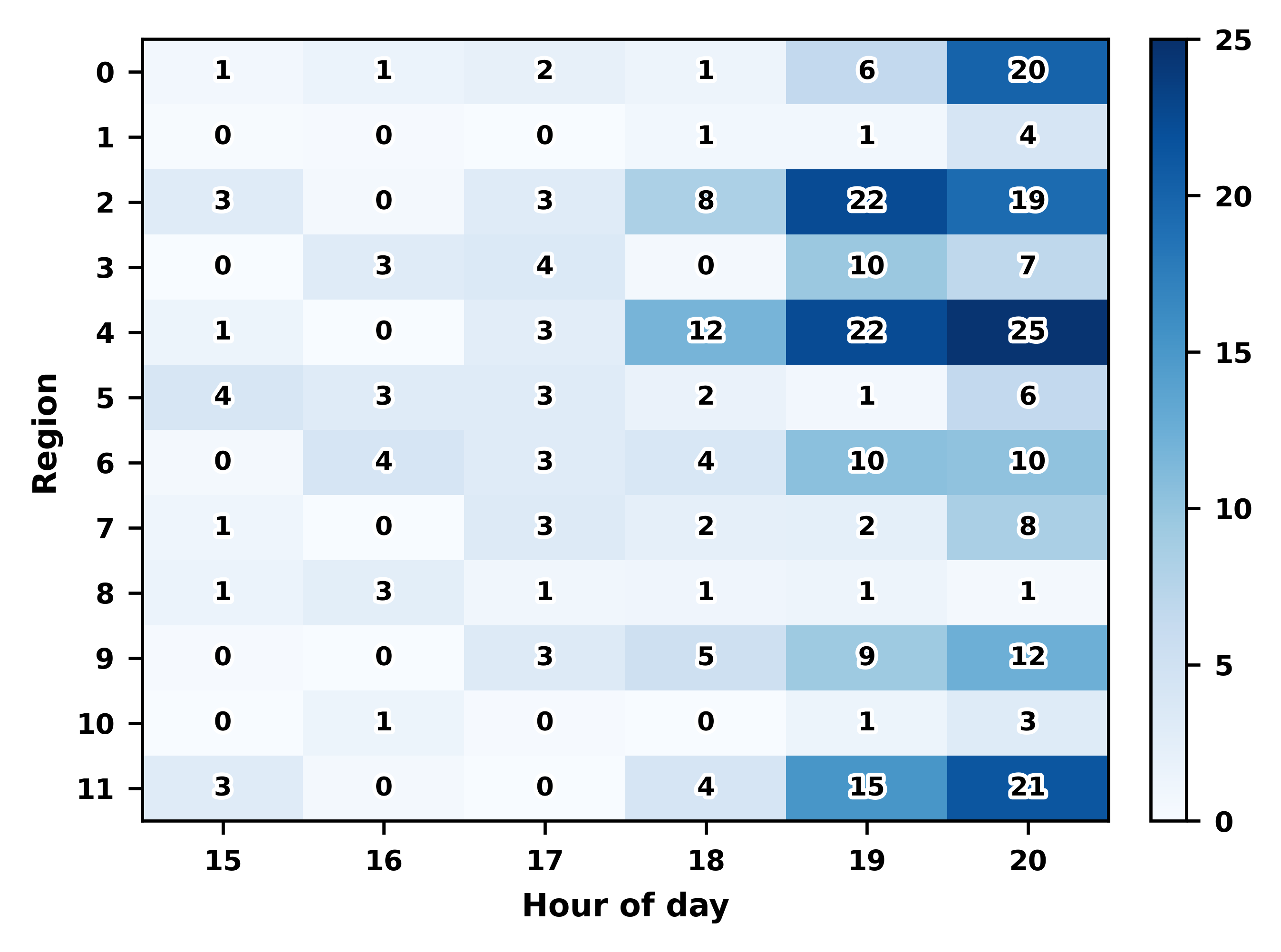}%
    }\hfill
    \subfloat[Penalty level (US\$)\label{fig:spatial_penaltylevel_reg}]{%
        \includegraphics[width=0.32\linewidth]{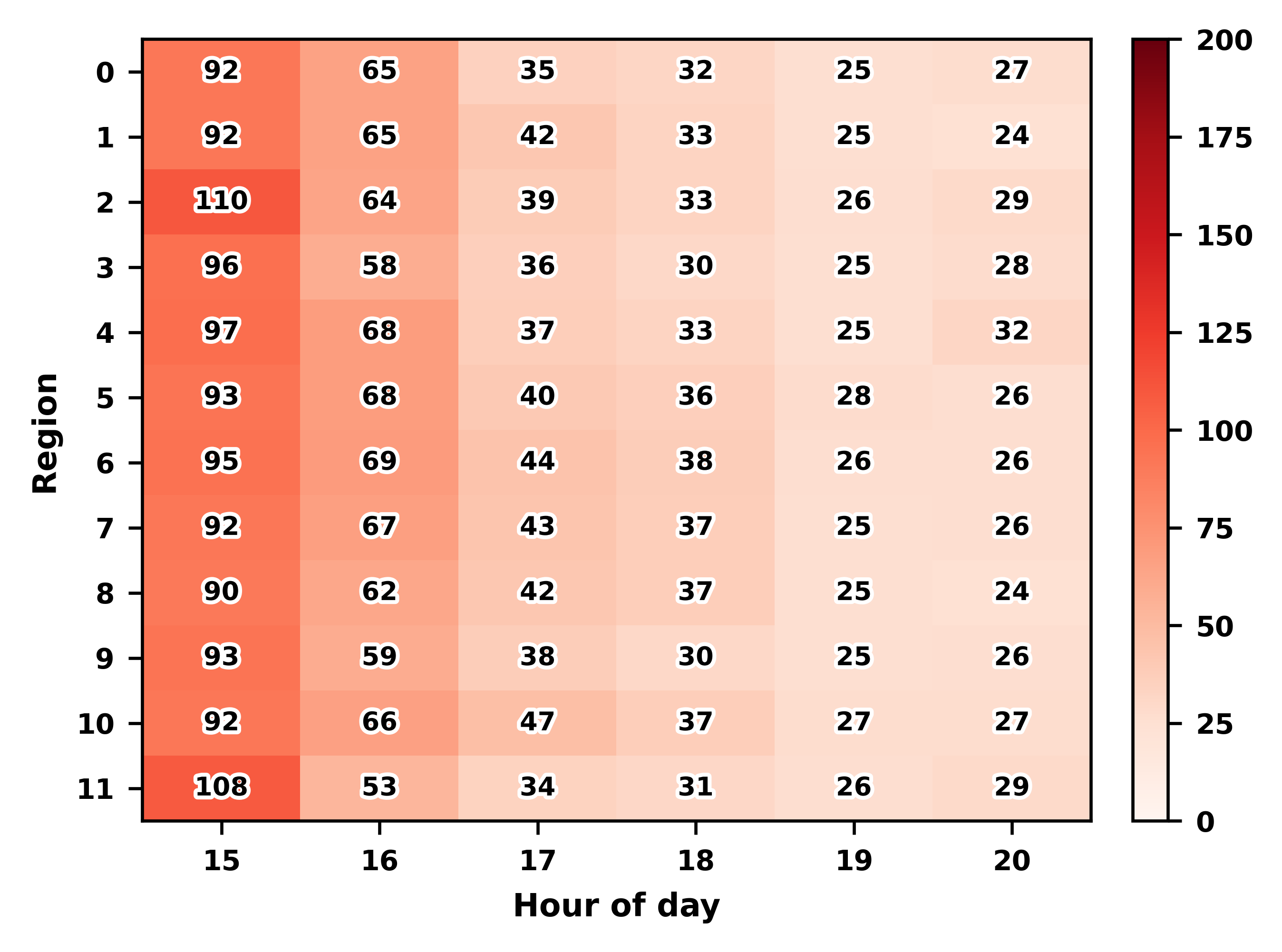}%
    }\hfill
    \subfloat[Penalty payment (US\$)\label{fig:spatial_penaltypaid_reg}]{%
        \includegraphics[width=0.32\linewidth]{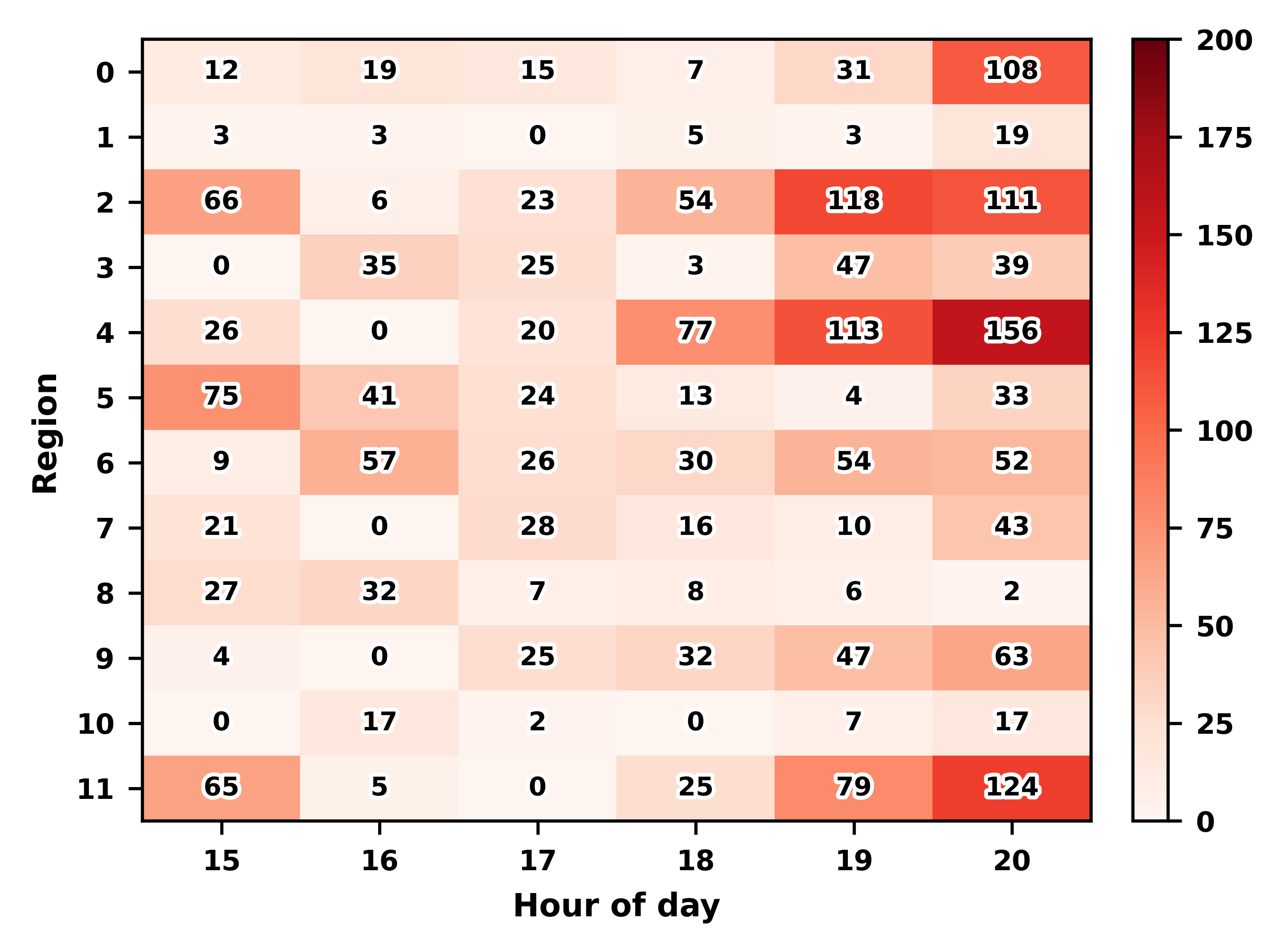}%
    }
    \caption[Regulator outcomes and policy under dynamic regional service availability regulation]{Regulator outcomes and policy with dynamic regional service availability regulation by region and hour in one representative evaluation run.}
    \label{fig:spatial_regulation_outcomes}
\end{figure*}

Dynamic regulation largely reduces afternoon deficits (Fig.~\ref{fig:spatial_regulation_outcomes}), while peak-period shortfalls persist. The learned penalty levels are highest in the early afternoon and lowest during the evening peak, with less pronounced regional variation. The regulator thus appears to limit intervention when further penalties offer little improvement. Penalty payments nevertheless peak in the evening because deficits remain large.

\begin{figure*}[!t]
    \centering

    \subfloat[Mean price factor, unregulated\label{fig:spatial_price_unreg}]{%
        \includegraphics[width=0.32\linewidth]{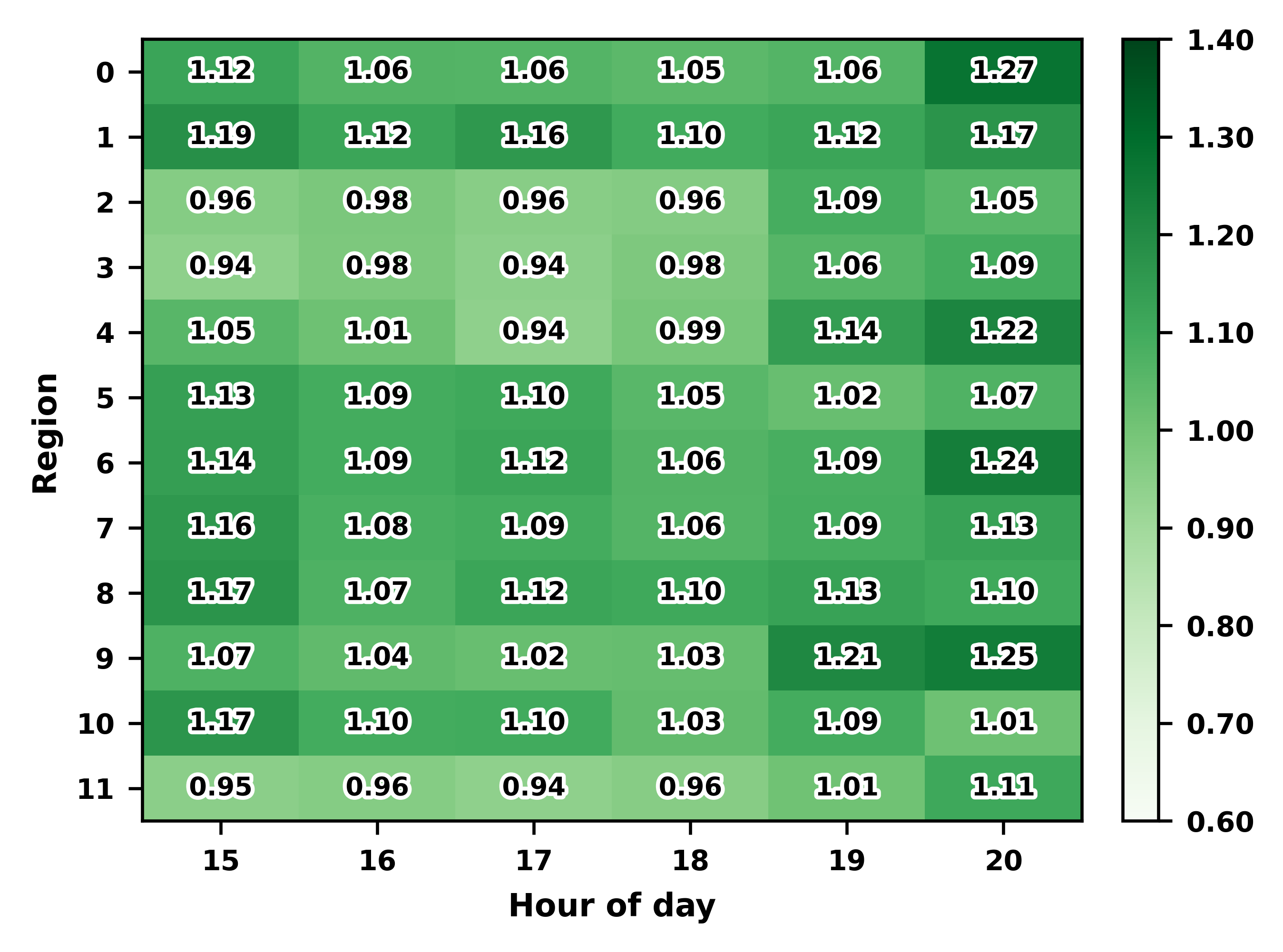}%
    }\hfill
    \subfloat[Rebalancing net flow (\#), unregulated\label{fig:spatial_reb_unreg}]{%
        \includegraphics[width=0.32\linewidth]{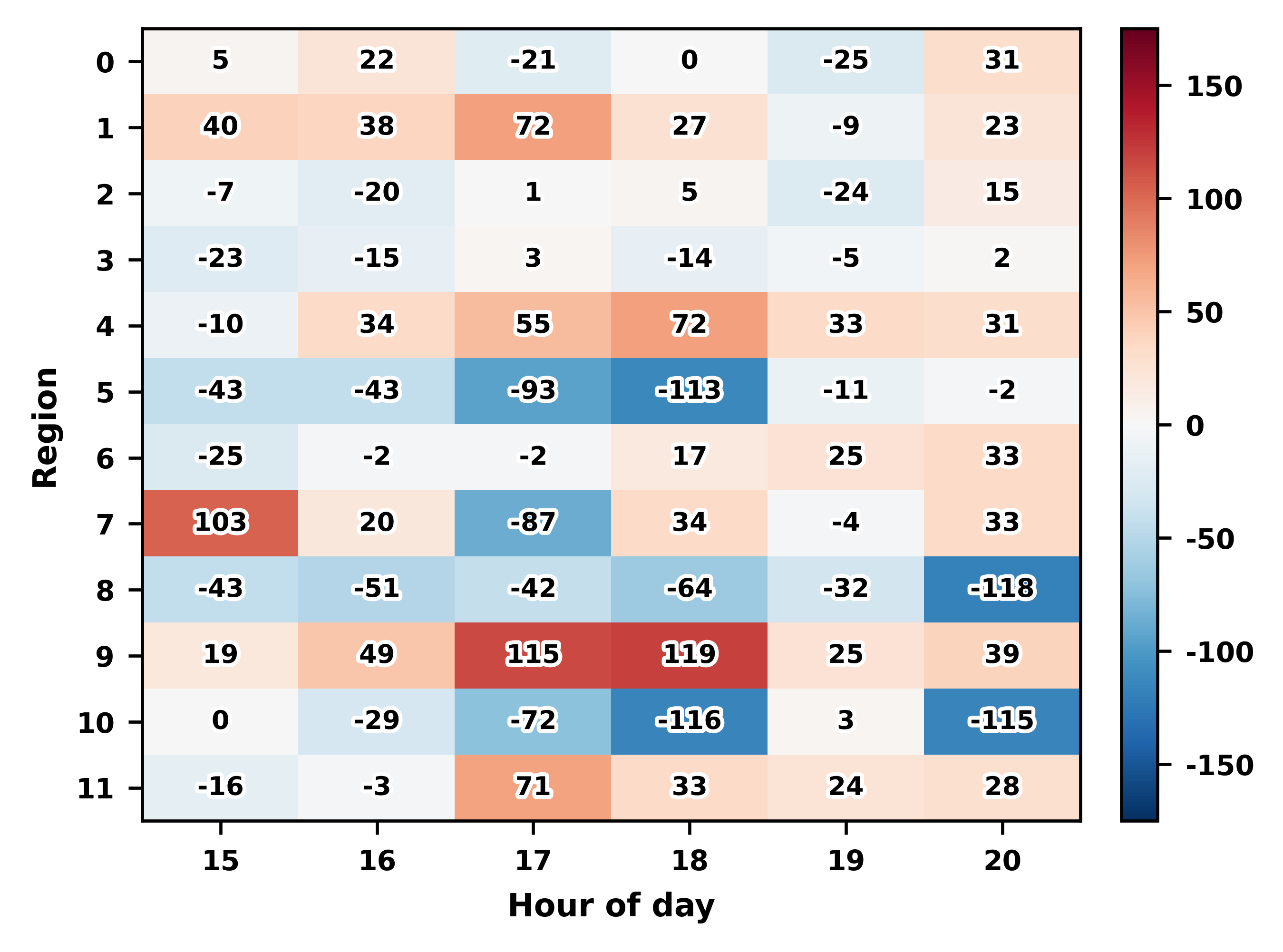}%
    }\hfill
    \subfloat[Realized utility per traveler (US\$), unregulated\label{fig:spatial_utility_unreg}]{%
        \includegraphics[width=0.32\linewidth]{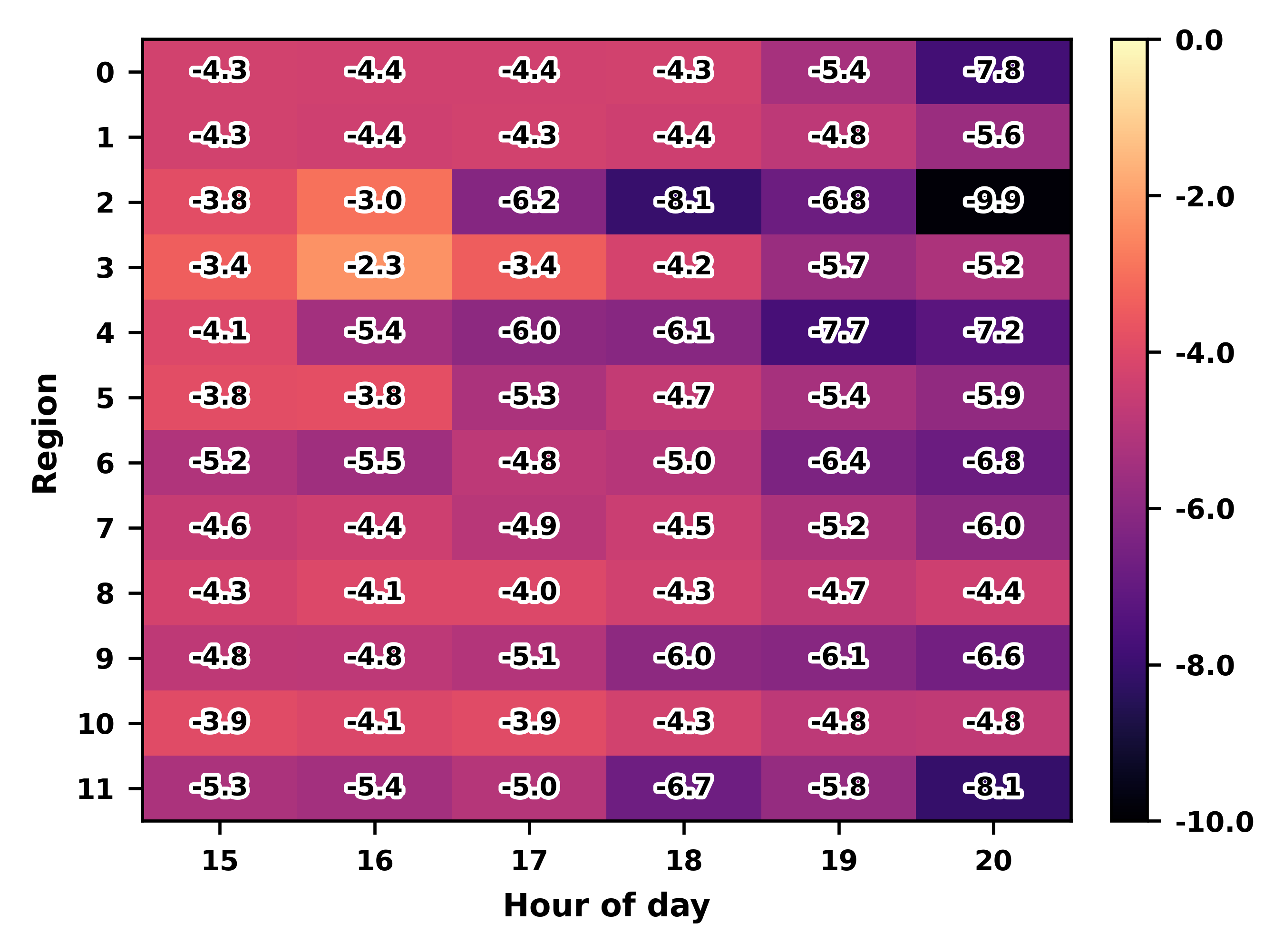}%
    }

    \vspace{-1em}

    \subfloat[Mean price factor, regulated\label{fig:spatial_price_reg}]{%
        \includegraphics[width=0.32\linewidth]{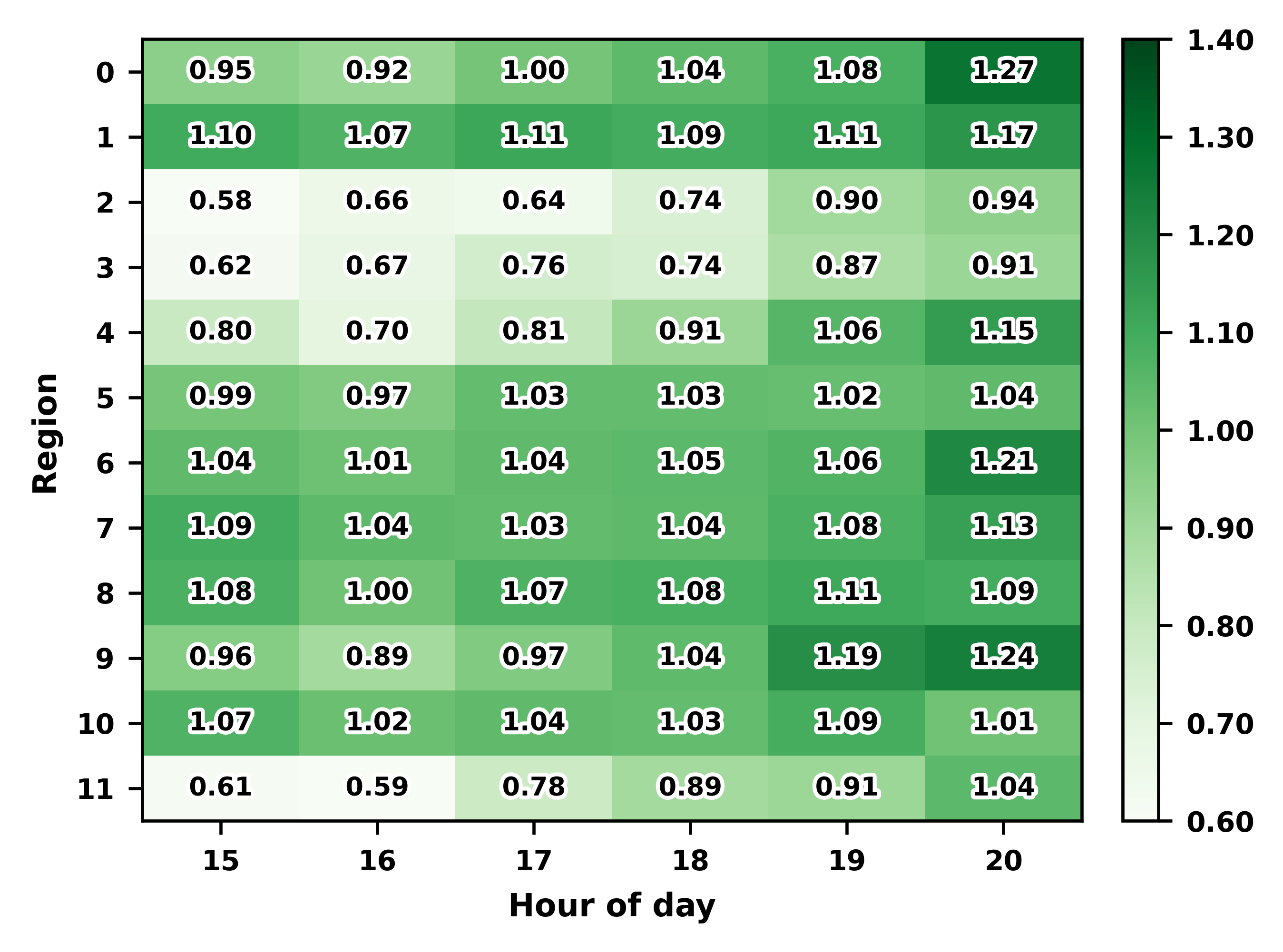}%
    }\hfill
    \subfloat[Rebalancing net flow (\#), regulated\label{fig:spatial_reb_reg}]{%
        \includegraphics[width=0.32\linewidth]{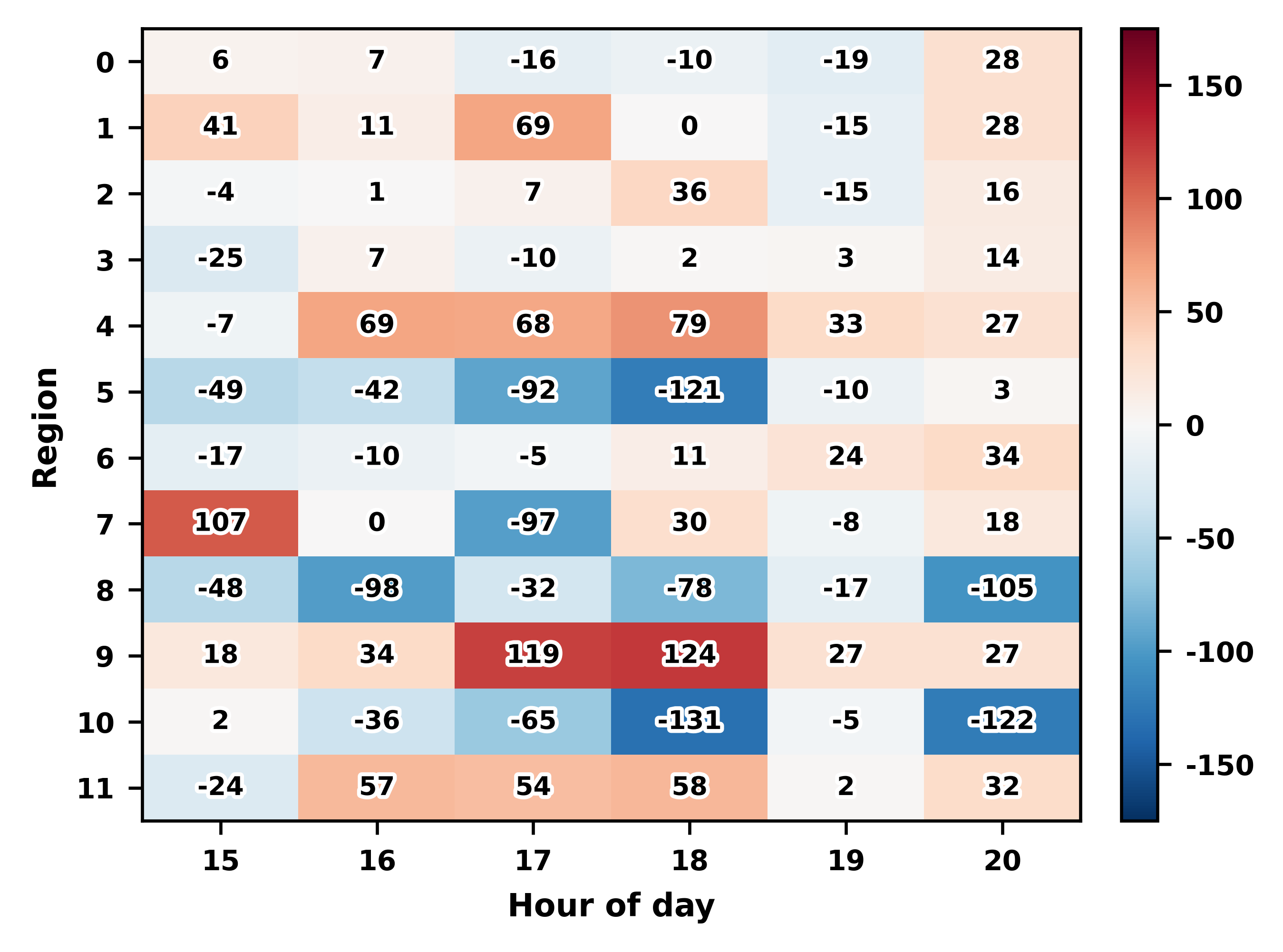}%
    }\hfill
    \subfloat[Realized utility per traveler (US\$), regulated\label{fig:spatial_utility_reg}]{%
        \includegraphics[width=0.32\linewidth]{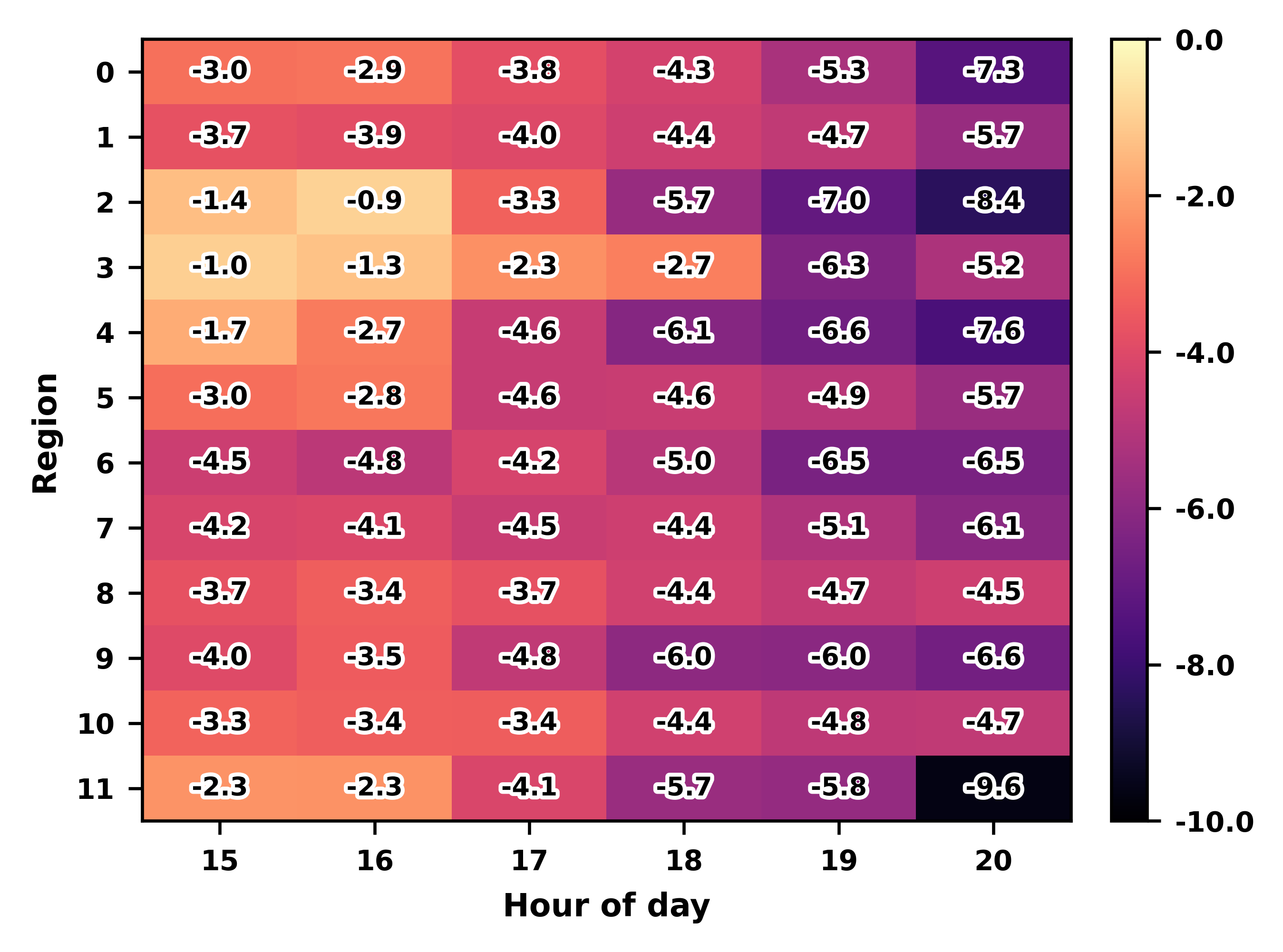}%
    }

    \caption[Operator policy and traveler outcomes with and without dynamic regional service availability regulation]{Operator policy and traveler outcomes with and without dynamic regional service availability regulation by region and hour in one representative evaluation run.}
    \label{fig:spatial_policy_response}
\end{figure*}

The operator lowers afternoon prices particularly in southern regions and redirects rebalancing toward them, improving service and traveler utility (Fig.~\ref{fig:spatial_policy_response}). Peak-period prices remain broadly stable, with only slight reductions in southern regions, and aggregate net rebalancing inflow to these regions increases only by eight vehicles. Deficits worsen in region 4 but improve in regions 2 and 3, suggesting redistribution of scarce capacity rather than elimination of the shortage. Traveler utility gains are correspondingly strongest in the afternoon and southern regions, whereas changes in the final hour are small and mixed. Monetary penalties therefore improve availability when operational flexibility exists but cannot guarantee compliance under peak demand.

\subsubsection{Sensitivity Analysis} \label{sec:results_spatial_equity_sensitivity}
Table~\ref{tab:spatial_sensitivity} shows the results for variations in fleet size, regulator decision interval, and target service rate under $\lambda=0.1$.


\begin{table*}[!ht]
\centering
\caption{Sensitivity analyses under dynamic spatial equity regulation.}
\label{tab:spatial_sensitivity}
\footnotesize
\setlength{\tabcolsep}{2pt}

\begin{tabular}{@{}l*{9}{r}@{}}
\toprule
& \multicolumn{3}{c}{\textbf{Fleet size} $N^{\mathrm{AMOD}}$}
& \multicolumn{3}{c}{\textbf{Regulator decision interval} $\Delta^{\mathrm{REG}}$}
& \multicolumn{3}{c}{\textbf{Target service rate} $s^{\mathrm{AMOD},*}$} \\
\cmidrule(lr){2-4}
\cmidrule(lr){5-7}
\cmidrule(lr){8-10}
\textbf{Metric}
& \textbf{850} & \textbf{1050} & \textbf{1250}
& \textbf{20} & \textbf{40} & \textbf{60}
& \textbf{0.6} & \textbf{0.7} & \textbf{0.8} \\
\midrule

\multicolumn{10}{@{}l}{\textbf{Regulator}} \\

Reward
& -124 (5) & -86 (5) & -72 (6)
& -124 (5) & -134 (8) & -138 (5)
& -124 (5) & -155 (7) & -202 (6) \\

Avg. penalty level (US\$)
& 22 (0.34) & 14 (0.21) & 22 (0.65)
& 22 (0.34) & 17 (0.34) & 43 (0.41)
& 22 (0.34) & 25 (0.95) & 36 (0.80) \\

Avg. service rate (\%)
& 78 (0) & 82 (0) & 85 (0)
& 78 (0) & 77 (1) & 79 (0)
& 78 (0) & 80 (0) & 80 (0) \\

Avg. service deficit (\%)
& 5 (0) & 3 (0) & 2 (0)
& 5 (0) & 5 (0) & 4 (0)
& 5 (0) & 7 (0) & 10 (0) \\

Affected regions (\%)
& 19 (1) & 14 (1) & 9 (1)
& 19 (1) & 21 (1) & 19 (0)
& 19 (1) & 26 (1) & 37 (1) \\

Penalties (US\$)
& 1,114 (79) & 622 (52) & 811 (138)
& 1,114 (79) & 939 (74) & 1,438 (119)
& 1,114 (79) & 1,541 (119) & 3,623 (152) \\

\midrule
\multicolumn{10}{@{}l}{\textbf{Operator}} \\

Reward (k US\$)
& 124.0 (0.7) & 124.9 (0.9) & 127.2 (1.0)
& 124.0 (0.7) & 125.7 (0.7) & 123.3 (0.6)
& 124.0 (0.7) & 121.1 (0.8) & 120.8 (0.9) \\

Rebalancing trips (\#)
& 2,128 (61) & 2,592 (66) & 2,526 (62)
& 2,128 (61) & 2,093 (60) & 2,194 (68)
& 2,128 (61) & 2,252 (78) & 2,242 (72) \\

Avg. price factor
& 0.98 (0.00) & 0.92 (0.00) & 0.91 (0.00)
& 0.98 (0.00) & 1.00 (0.00) & 0.96 (0.00)
& 0.98 (0.00) & 0.92 (0.00) & 0.95 (0.00) \\

Served demand (\#)
& 16,584 (69) & 17,997 (101) & 18,656 (94)
& 16,584 (69) & 16,645 (71) & 16,800 (77)
& 16,584 (69) & 17,061 (74) & 17,023 (84) \\

Lost demand (\#)
& 418 (45) & 249 (30) & 97 (27)
& 418 (45) & 477 (43) & 379 (31)
& 418 (45) & 697 (51) & 563 (31) \\

Avg. wait (min)
& 0.51 (0.03) & 0.36 (0.03) & 0.17 (0.03)
& 0.51 (0.03) & 0.60 (0.04) & 0.55 (0.04)
& 0.51 (0.03) & 0.68 (0.04) & 0.63 (0.04) \\

Vehicle utilization (\%)
& 67 (0) & 59 (0) & 51 (0)
& 67 (0) & 67 (0) & 68 (1)
& 67 (0) & 70 (0) & 69 (0) \\

\midrule
\multicolumn{10}{@{}l}{\textbf{Traveler}} \\

Realized utility (k US\$)
& -109.5 (0.9) & -95.3 (0.8) & -90.2 (1.0)
& -109.5 (0.9) & -110.4 (0.9) & -107.3 (0.9)
& -109.5 (0.9) & -103.9 (1.0) & -105.2 (0.9) \\

AMoD mode share (\%)
& 76 (0) & 81 (0) & 84 (0)
& 76 (0) & 76 (0) & 77 (0)
& 76 (0) & 79 (0) & 79 (0) \\

\bottomrule
\end{tabular}

\vspace{0.4em}
\begin{minipage}{\linewidth}
\footnotesize
\end{minipage}
\end{table*}
\textit{Fleet size:} Increasing the fleet improves regulator reward from $-124$ to $-72$ through reduced service deficits and fewer affected regions without higher mean penalty levels. This supports a capacity-related explanation for remaining deficits in the baseline. Lower prices and more rebalancing increase served demand and operator reward, but utilization and reward per vehicle fall. Larger fleets therefore improve availability at the expense of fleet productivity and shift demand from \gls{acr:pt}.

\textit{Regulator decision interval:} Extending $\Delta^{\mathrm{REG}}$ from 20 to 60 time steps (one to three hours) lowers regulator reward from $-124$ to $-138$, although all tested intervals exceed the hourly policy's matched static benchmark of $-147$ (Table~\ref{tab:spatial_comparison}). Availability outcomes do not follow the same monotonic pattern as regulator reward. The three-hour setting achieves lower deficits than hourly regulation but imposes higher mean penalties, whereas the two-hour setting has more affected regions but lower mean penalties. This suggests that sufficiently frequent data access matters for regulatory adaptivity. 

\textit{Target service rate:} Raising the target increases mean penalties from US\$22 to US\$36 and payments from US\$1,114 to US\$3,623, reducing operator reward. Regulator outcomes are not directly comparable because their definitions depend on the target. Prices are lowest, rebalancing most intensive, and traveler utility highest at the intermediate 70\% target. At 80\%, the operator no longer intensifies its response. Targets should therefore reflect operating constraints, as greater ambition can increase operator costs without further traveler benefits.


\section{Conclusion} \label{chap:conclusion}

We proposed a closed-loop formulation of \gls{acr:amod} regulation in which a regulator acts on a learning-based operator through penalties at a slower time scale. Static (open-loop) regulation is formulated as an \gls{acr:mdp} for the operator, and dynamic (closed-loop) regulation as a \gls{acr:posg} in which a partially observing regulator adjusts region-specific penalties to induce compliance with minimal penalties. The operator's joint pricing and rebalancing policy and the regulator's penalty policy are learned with graph-based (multi-agent) \gls{acr:rl} and evaluated on real-world Manhattan data.

For surge pricing, static regulation is sufficient to nearly eliminate excessive prices, reducing the share of affected travelers from 22\% to 1\%. Average traveler outcomes improve, although some travelers are worse off at peak periods, when surge pricing helps ration limited capacity. For regional service availability, dynamic regulation improves regulator reward by 10.5--15.6\% over static benchmarks with matching average penalty levels at intermediate reward weights. Higher penalties reduce service deficits and affected regions, but marginal availability gains diminish and peak-period deficits persist. The learned regulator responds by reducing penalties when they no longer improve compliance. Both instruments can improve their targeted outcomes but place financial and operational pressure on the operator and shift demand from \gls{acr:pt} to \gls{acr:amod}, highlighting the need to balance instrument-specific targets with broader mobility objectives.

For practice, these results suggest that excessive surge pricing can be addressed with simple fixed penalties on prices above a threshold, without adaptive regulation. Regional availability targets, in contrast, are bounded by fleet capacity, where adaptive penalties improve availability when spare capacity exists but cannot enforce compliance under peak demand, so targets should reflect operating constraints. The benefit of adaptive penalties declines as regulatory updates become less frequent, although even three-hourly updates outperform matched fixed penalties, so timely access to operational data increases the value of dynamic regulation. 

The observed trade-offs motivate future research on broader regulatory objectives, such as traveler utility, and coordinated interventions in \gls{acr:amod} and \gls{acr:pt}, including dynamic fare subsidies as proposed by \cite{kadem2026dynamicincentivization}. Further research could test the effectiveness of regulation under more realistic operator and traveler behavior by introducing competing operators and heterogeneous traveler populations \cite{toft2026competitive}, including travelers without a viable \gls{acr:pt} alternative, for whom availability targets could be defined directly, and incorporating expected \gls{acr:amod} service quality into mode choice. These extensions could alter responses to regulatory incentives and the penalty levels required to achieve regulatory objectives.


\bibliographystyle{IEEEtran}
\bibliography{literature} 

{\appendix
\section*{Vehicle Rebalancing Model} \label{app:vrm}
Following \cite{gammelli2021gnnrl}, the rebalancing model converts the operator's desired idle-vehicle distribution into minimum-cost rebalancing flows $y_{i,j,t}^{\mathrm{AMOD}}$ from region $i$ to $j$:

\begin{align*}
\min \quad 
& \sum_{(i,j)\in\mathcal{E}} c_{i,j,t}^{\mathrm{AMOD}} \cdot y_{i,j,t}^{\mathrm{AMOD}} \\ 
\mathrm{s.t.} \quad
& \sum_{j \neq i} \left( y_{j,i,t}^{\mathrm{AMOD}} - y_{i,j,t}^{\mathrm{AMOD}} \right) + n_{i,t}^{\mathrm{AMOD}}
\geq \tilde{n}_{i,t}^{\mathrm{AMOD}},
\quad i \in \mathcal{V} \\
& \sum_{j \neq i} y_{i,j,t}^{\mathrm{AMOD}}
\leq n_{i,t}^{\mathrm{AMOD}},
\quad i \in \mathcal{V} \\
& y_{i,j,t}^{\mathrm{AMOD}} \geq 0,
\quad (i,j) \in \mathcal{E} 
\end{align*}
where the target vehicle counts are derived from the operator policy output as $\tilde{n}_{i,t}^{\mathrm{AMOD}}=\left\lfloor w_{i,t}^{\mathrm{AMOD}}\sum_{j\in\mathcal{V}}n_{j,t}^{\mathrm{AMOD}}\right\rfloor$.

\vfill

\end{document}


\twocolumn
\title{Supplementary Materials}



\maketitle


\newacronym{acr:amod}{AMoD}{autonomous Mobility-on-Demand}
\newacronym{acr:mod}{MoD}{Mobility-on-Demand}
\newacronym{acr:pt}{PT}{public transport}
\newacronym{acr:msp}{MSP}{mobility service provider}
\newacronym[longplural={municipal transportation authorities}]{acr:mta}{MTA}{municipal transportation authority}

\newacronym{acr:mpc}{MPC}{Model Predictive Control}
\newacronym{acr:ppo}{PPO}{Proximal Policy Optimization}

\newacronym{acr:mdp}{MDP}{Markov Decision Process}
\newacronym{acr:posg}{POSG}{Partially Observable Stochastic Game}

\newacronym{acr:drl}{DRL}{deep reinforcement learning}
\newacronym{acr:rl}{RL}{reinforcement learning}

\newacronym{acr:sac}{SAC}{Soft Actor-Critic}
\newacronym{acr:a2c}{A2C}{Advantage Actor-Critic}
\newacronym{acr:td}{TD}{temporal-difference}

\newacronym{acr:gcn}{GCN}{graph convolutional network}
\newacronym{acr:fc}{FC}{fully connected}
\newacronym{acr:lrelu}{LeakyReLU}{Leaky Rectified Linear Unit}
\newacronym{acr:relu}{ReLU}{Rectified Linear Unit}

\newacronym{acr:fcfs}{FCFS}{first-come, first-served}
\newacronym{acr:od}{OD}{origin-destination}

\newacronym{acr:gtfs}{GTFS}{General Transit Feed Specification}



\section{Model Architecture} \label{app:architecture}

\begin{strip}
    \centering
    \includegraphics[width=\textwidth]{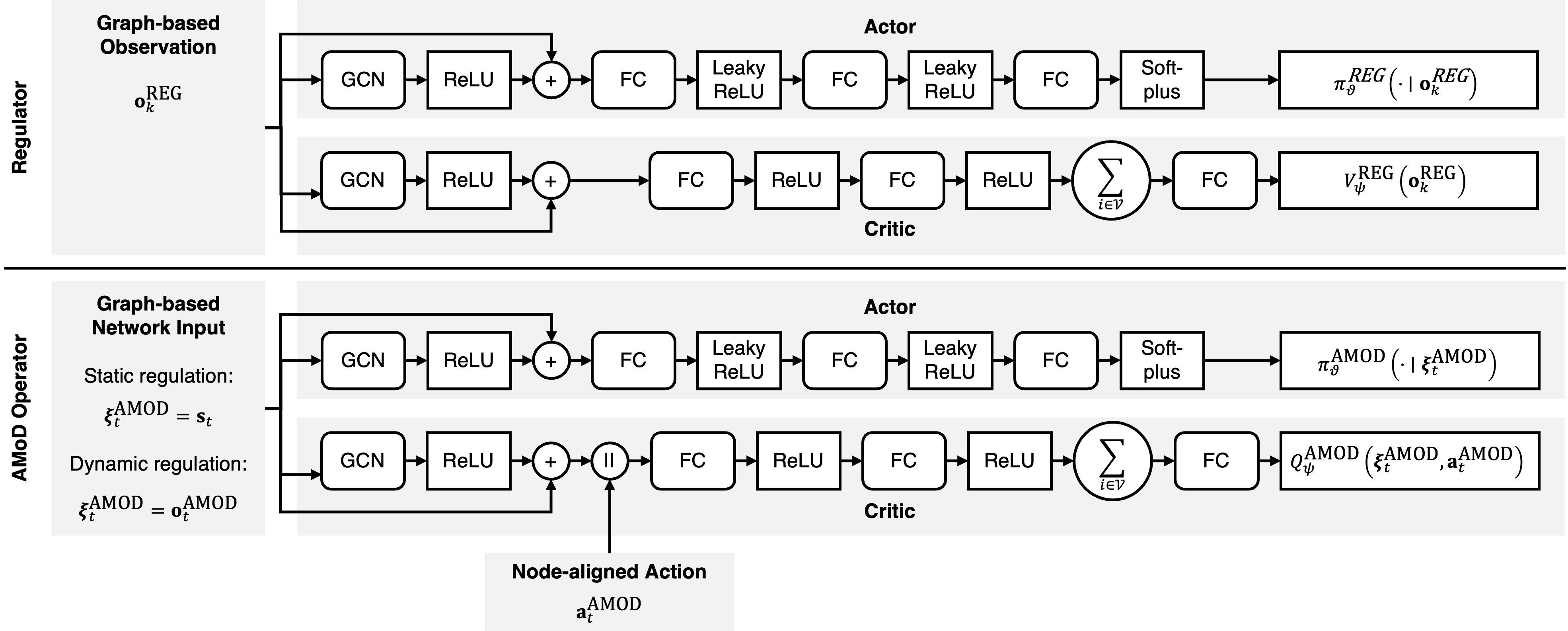}
    \captionof{figure}{Regulator and operator model architecture under
                       static and dynamic regulation}
    \label{fig:model_architecture}
\end{strip}

  \section{Hyperparameters Used in Training} \label{app:hyperparameters}

\begin{table}[!ht]
\centering
\caption{Hyperparameters used for operator and regulator training.}
\label{tab:hyperparameters}
\footnotesize
\setlength{\tabcolsep}{3pt}
\begin{tabular}{lcccc}
\toprule
&
\multicolumn{3}{c}{\textbf{Operator}} &
\multicolumn{1}{c}{\textbf{Regulator}} \\
\cmidrule(lr){2-4} \cmidrule(lr){5-5}
&
\multicolumn{2}{c}{Static} &
\multicolumn{1}{c}{Dynamic} &
\multicolumn{1}{c}{Dynamic} \\
\cmidrule(lr){2-3} \cmidrule(lr){4-4} \cmidrule(lr){5-5}
\textbf{Hyperparameter}
& SA
& SP
& SA
& SA \\
\midrule

\multicolumn{5}{l}{\textit{Training}} \\

Algorithm
& SAC & SAC & SAC & A2C \\

Initialization strategy
& WS
& FS
& WS
& FS \\

Training episodes
& $20{,}000$
& $20{,}000$
& $20{,}000$
& $20{,}000$ \\

Critic warm-up episodes
& --
& --
& --
& $50$ \\

Actor learning rate
& $1 \times 10^{-3}$
& $1 \times 10^{-3}$
& $1 \times 10^{-3}$
& $2 \times 10^{-4}$ \\

Critic learning rate
& $1 \times 10^{-3}$
& $1 \times 10^{-3}$
& $1 \times 10^{-3}$
& $6 \times 10^{-4}$ \\

Actor gradient clipping
& $10$
& $10$
& $10$
& $1000$ \\

Critic gradient clipping
& $500$
& $500$
& $500$
& $1000$ \\

Reward scaling factor
& $0.01$
& $0.01$
& $0.01$
& $1.0$ \\

Discount factor $\gamma$
& --
& --
& --
& $0.97$ \\

Entropy coefficient $\alpha$
& $0.1$
& $0.1$
& $0.1$
& -- \\

Batch size
& $100$
& $100$
& $100$
& -- \\

Polyak coefficient
& $0.995$
& $0.995$
& $0.995$
& -- \\

\midrule

\multicolumn{5}{l}{\textit{Network}} \\

Hidden layer size $h$
& $256$
& $256$
& $256$
& $256$ \\

Look-ahead horizon $H^\mathrm{P}$
& $6$
& $6$
& $6$
& $6$ \\

Input scaling factor
& $0.01$
& $0.01$
& $0.01$
& -- \\

\bottomrule
\end{tabular}

\vspace{0.5em}
\begin{minipage}{\linewidth}
\footnotesize
\textit{Notes:} SA = regional service availability; SP = surge pricing; WS = Warm-start; FS = From scratch.
\end{minipage}

\end{table}

\section{Scenario Data and Its Construction} \label{app:data}
The operator manages 850 vehicles at an operating cost of US\$18 per vehicle-hour \cite{tresca2025robotaxicoordination}. Travelers have an hourly wage of US\$22.77 \cite{toft2026competitive} and a maximum waiting time of six minutes. AMoD and PT data inputs for the evaluation scenario are constructed as described below.

\subsection*{Taxi-Based AMoD Inputs}
The historical reference demand $\bar{d}_{i,j,t}$, AMoD trip durations $\tau_{i,j,t}^{\mathrm{AMOD}}$, and AMoD reference prices $\bar{p}_{i,j,t}^{\mathrm{AMOD}}$ are constructed from historical New York City taxi trip records provided by \cite{illinoisdatabank}, using the preprocessing approach of \cite{toft2026competitive}. The dataset contains individual taxi trips between 2010 and 2013, including origin and destination locations, trip durations, and fare information.

Taxi records are assigned to ZIP-code regions based on their origin and destination coordinates, filtered to the spatial area, day, and time window of the evaluation scenario, and assigned to simulation time steps according to their departure time. Trips with identical origin and destination regions are removed. For each \gls{acr:od} pair $(i,j)$ and time step $t$, the historical reference demand $\bar{d}_{i,j,t}$ is given by the number of observed trips, while the AMoD reference price $\bar{p}_{i,j,t}^{\mathrm{AMOD}}$ and trip duration $\tau_{i,j,t}^{\mathrm{AMOD}}$ are computed as the corresponding mean fare and mean travel duration.

\subsection*{PT Inputs} 
Corresponding PT travel times $\tau_{i,j,t}^{\mathrm{PT}}$ are obtained using the FleetPy router \cite{Engelhardt.2026} and NYC bus and subway \gls{acr:gtfs} data \cite{mta_bus_gtfs_2013,mta_subway_gtfs_2013}, with a fare of US\$2.50 per trip \cite{fitzsimmons2015mtafares}. 

\textit{PT network construction:}
The PT network is constructed from New York City bus and subway \gls{acr:gtfs} data covering the scenario day. In addition to standard \gls{acr:gtfs} files, the router requires a station file that maps PT stops to stations and provides their geographical locations. For the bus network, each stop is treated as a separate station, implying zero transfer time between the stop and its assigned station. For the subway network, parent stations defined in the \gls{acr:gtfs} stop file are used as stations using the predefined transfer times. 

The \gls{acr:gtfs} transfer file is extended to include subway--bus, bus--subway, and bus--bus transfers whenever the corresponding stations are within a Manhattan distance of 400 meters. Transfer times are computed from the corresponding walking distance using an assumed walking speed of 5 kilometers per hour \cite{murtagh2021walkingspeed}.

\textit{Routing:}
Since the FleetPy router operates on PT stops rather than geographical coordinates, trip origins and destinations are first mapped to candidate access and egress stations. For each trip, bus stations within a Manhattan distance of 400 meters and subway stations within a Manhattan distance of 800 meters are considered candidate stations \cite{elgeneidy2014walkingdistances}. Access and egress walking times are then computed from the corresponding Manhattan distances using the assumed walking speed.

If the origin and destination candidate station sets overlap, the overlapping stations are removed from the PT routing candidates and an additional walking-only option is added. Its travel time is computed consistently with the access and egress walking times. For each trip, candidate stations are expanded to their associated stops, and the router is queried for all feasible \gls{acr:od} stop combinations. The selected PT connection is the fastest available route, where total travel duration includes in-vehicle time, transfer time, and access and egress walking times. If the walking-only option is faster than all routed PT alternatives, the walking option is selected instead.

Finally, PT travel durations are aggregated to the simulation resolution. For each \gls{acr:od} pair $(i,j)$ and time step $t$, $\tau_{i,j,t}^{\mathrm{PT}}$ is computed as the mean PT travel duration across all trips assigned to the corresponding aggregation cell.

\section{Scenario characteristics} \label{app:scenario_characteristics}

\begin{table}[!ht]
\centering
\caption[Hourly demand]{Total demand by hour of day.}
\label{tab:demand_by_hour}
\footnotesize

\begin{tabular}{lcccccc}
\toprule
\textbf{Hour of day} 
& 15 & 16 & 17 & 18 & 19 & 20 \\

\midrule

\textbf{Mean}      
& 3,793 & 2,982 & 4,346 & 7,091 & 9,069 & 10,021 \\

\textbf{Std. dev.}
& 67 & 54 & 57 & 84 & 88 & 92\\
\bottomrule
\end{tabular}

\vspace{0.5em}
\begin{minipage}{\linewidth}
\footnotesize
\centering
\textit{Notes:} Values are computed over 100 simulation samples.
\end{minipage}

\end{table}

\begin{figure}[H]
    \centering

    \subfloat[Mean demand share (\%) by hour and origin region%
        \label{fig:results_demandshare}]{%
        \includegraphics[width=\linewidth]{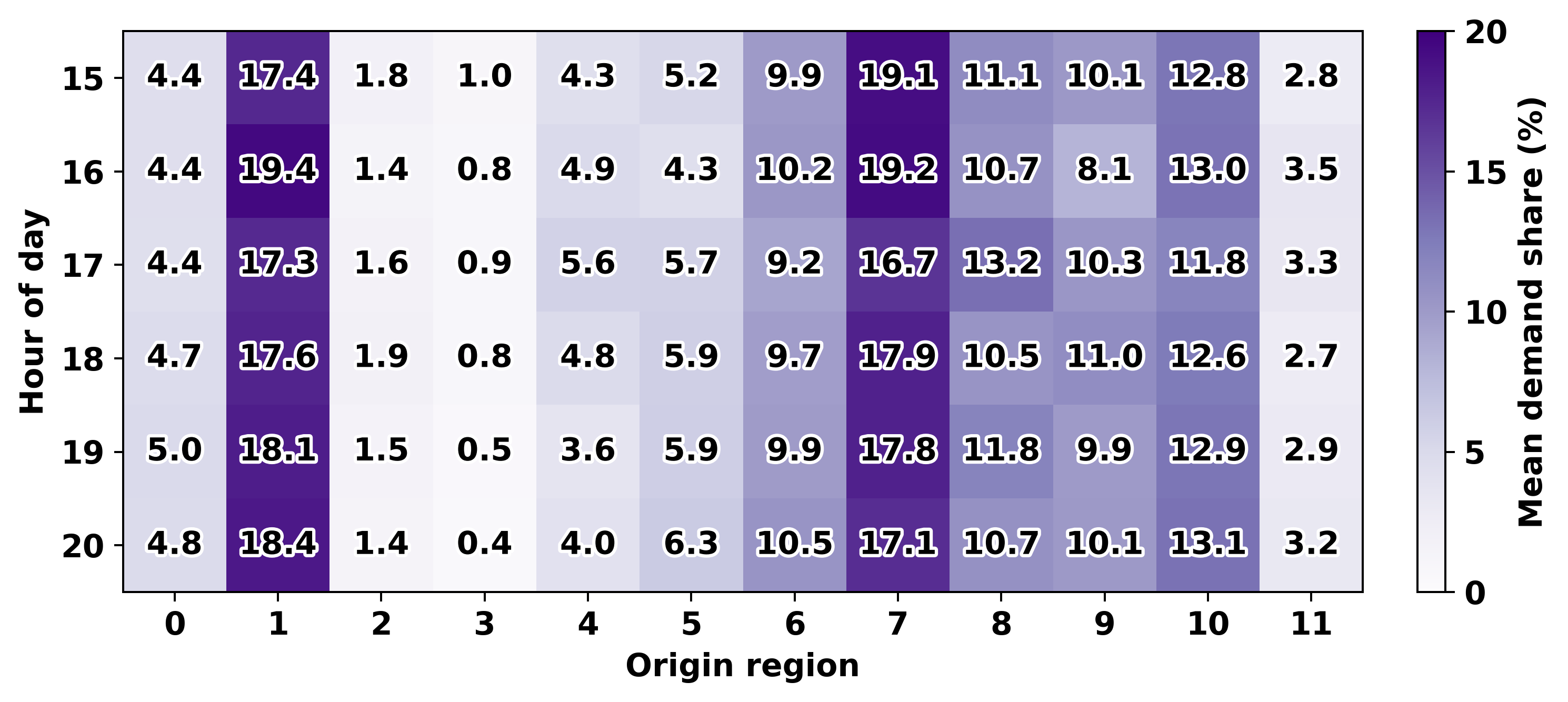}%
    }

    \vspace{-0.5em}

    \subfloat[Mean net demand flow (destination minus origin)%
        \label{fig:results_netdemandflow}]{%
        \includegraphics[width=\linewidth]{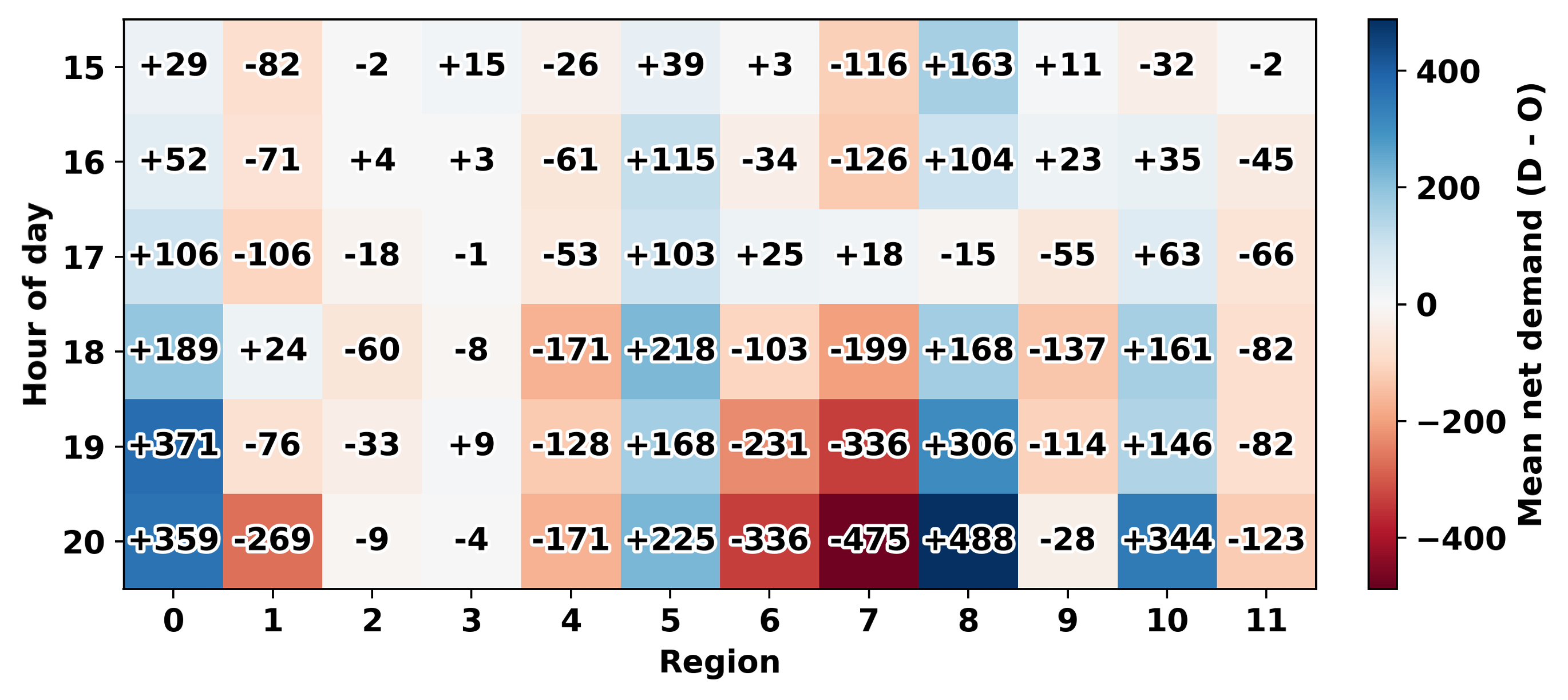}%
    }

    \vspace{-0.5em}

    \subfloat[Mean AMoD mode share (\%) by hour%
        \label{fig:results_amodsharehour}]{%
        \includegraphics[width=\linewidth]{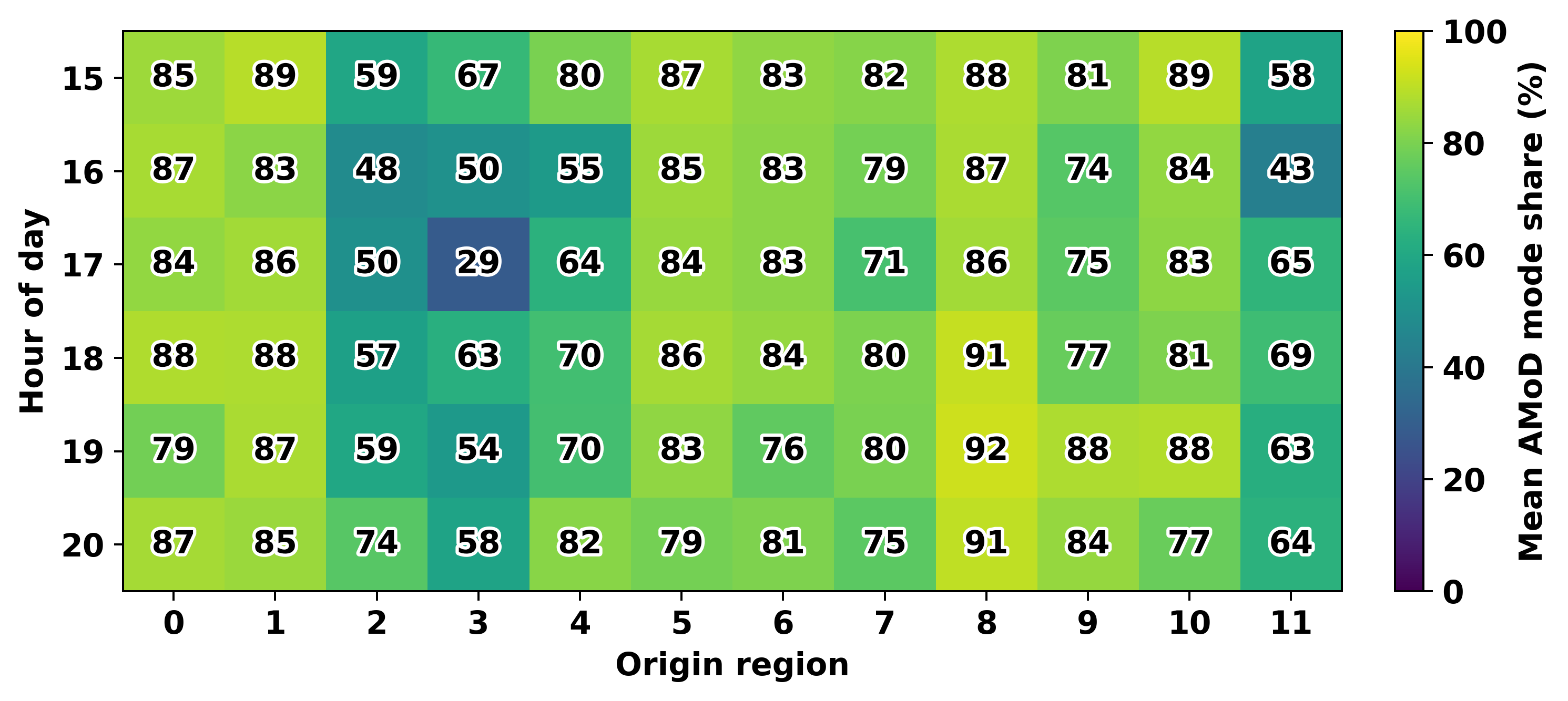}%
    }

    \vspace{-0.5em}

    \subfloat[Mean AMoD mode share (\%) by price factor%
        \label{fig:results_amodshareprice}]{%
        \includegraphics[width=\linewidth]{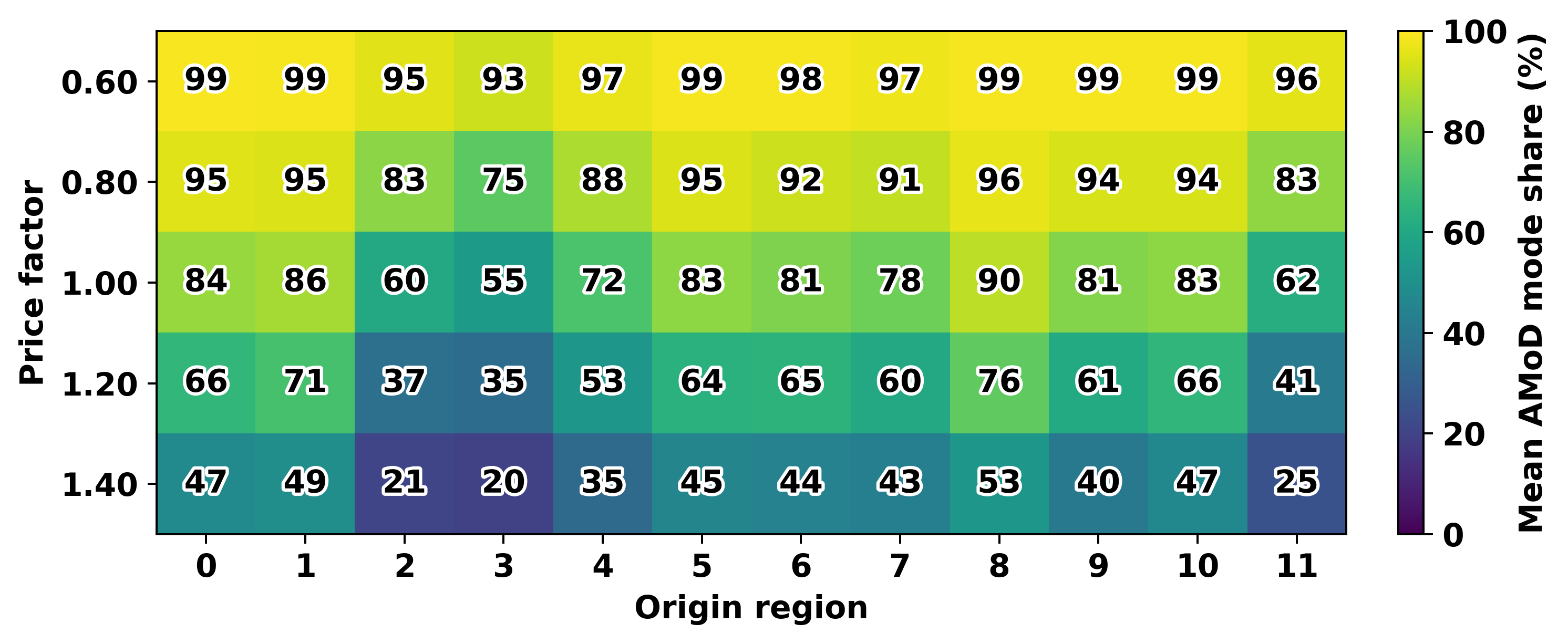}%
    }

    \captionsetup{skip=2pt}
    \caption{Scenario characteristics over 100 simulation samples.}
    \label{fig:results_demand_amod}
\end{figure}

\subsection*{Factors Influencing AMoD Mode Shares at Historical Reference Prices}
\label{app:amod_mode_share_factors}

\begin{figure}[H]
    \centering
    \includegraphics[width=\linewidth]{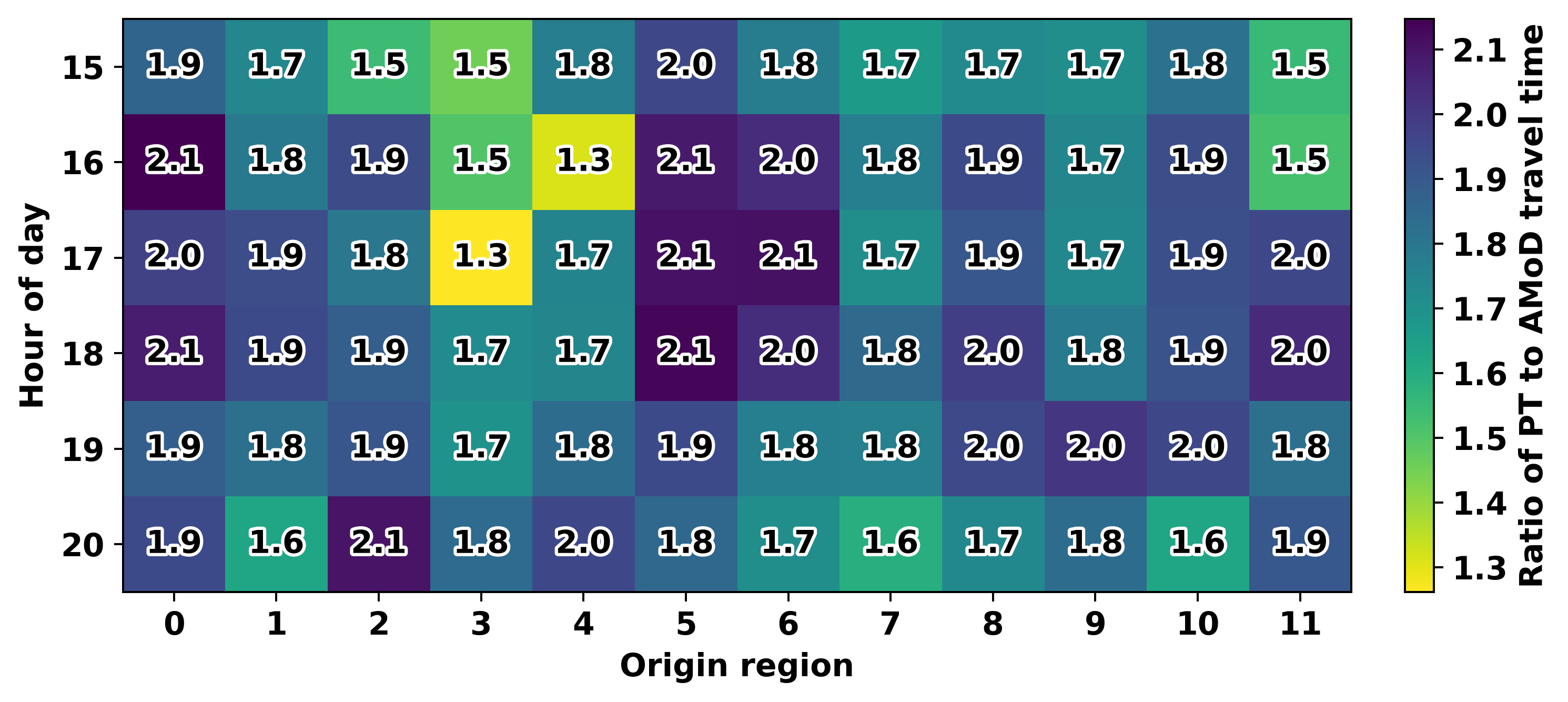}
    \caption{PT-to-AMoD travel-time ratio by origin region and hour, averaged over 100 simulation samples.}
    \label{app:fig_app_timeratio}
\end{figure}

\begin{figure}[H]
    \centering
    \includegraphics[width=\linewidth]{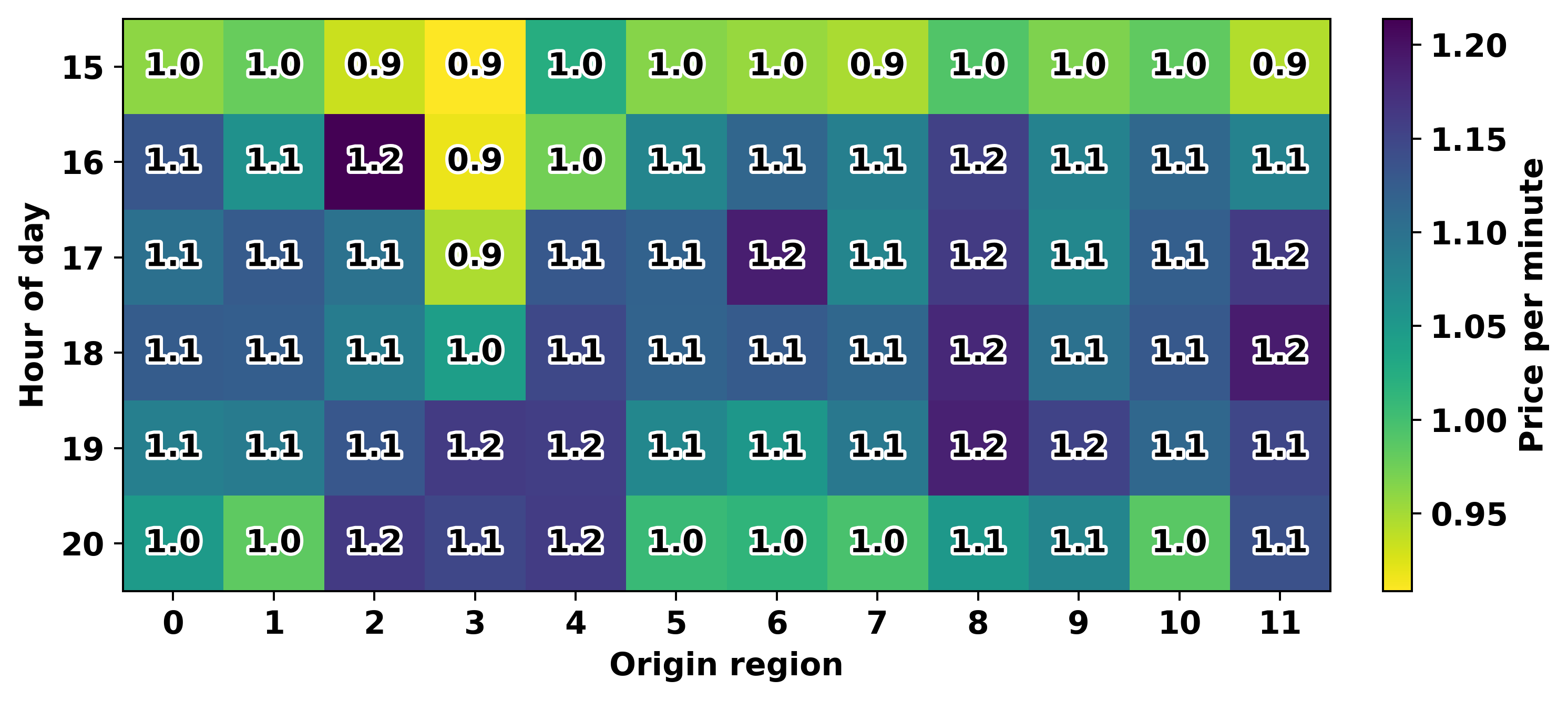}
    \caption{AMoD reference price in US\$ per minute of travel time by origin region and hour, averaged over 100 simulation samples.}
    \label{app:fig_app_pricepermin}
\end{figure}

Figure~\ref{app:fig_app_timeratio} reports the PT-to-AMoD travel-time ratio, and Figure~\ref{app:fig_app_pricepermin} reports the AMoD reference price per minute of travel time by region and hour. While AMoD has a relative travel-time disadvantage in some southern regions during the early afternoon, the PT-to-AMoD travel-time ratio in later hours is comparable to other regions. In the evening, however, AMoD reference prices per minute of travel time are comparatively high in southern regions. This suggests that the low baseline AMoD mode shares in these regions are not solely driven by superior PT travel times, but also by price-related conditions that can be influenced by the operator.

\section{Operator Sensitivity to Surge Pricing Regulation} \label{app:surge_sensitivity}

\begin{figure}[H]
\centering
\includegraphics[width=0.98\linewidth]{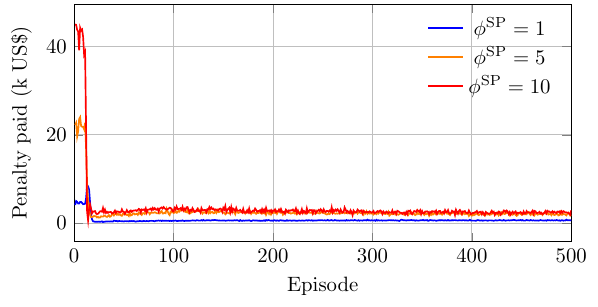}
\caption[Surge pricing penalties during training]{Surge pricing
penalties paid by the operator during training per episode for the
first 500 training episodes under surge pricing penalty levels
$\phi^{\mathrm{SP}} \in \{1,5,10\}$.}
\label{fig:surge_training_penalties}
\end{figure}

\begin{strip}
\begin{minipage}{\textwidth}

\section{Regional Performance Under Surge Pricing Regulation}

\begin{figure}[H]
    \centering
    \subfloat[Rebalancing net flow (\#), unregulated\label{fig:surge_reb_unreg}]{%
        \includegraphics[width=0.32\linewidth]{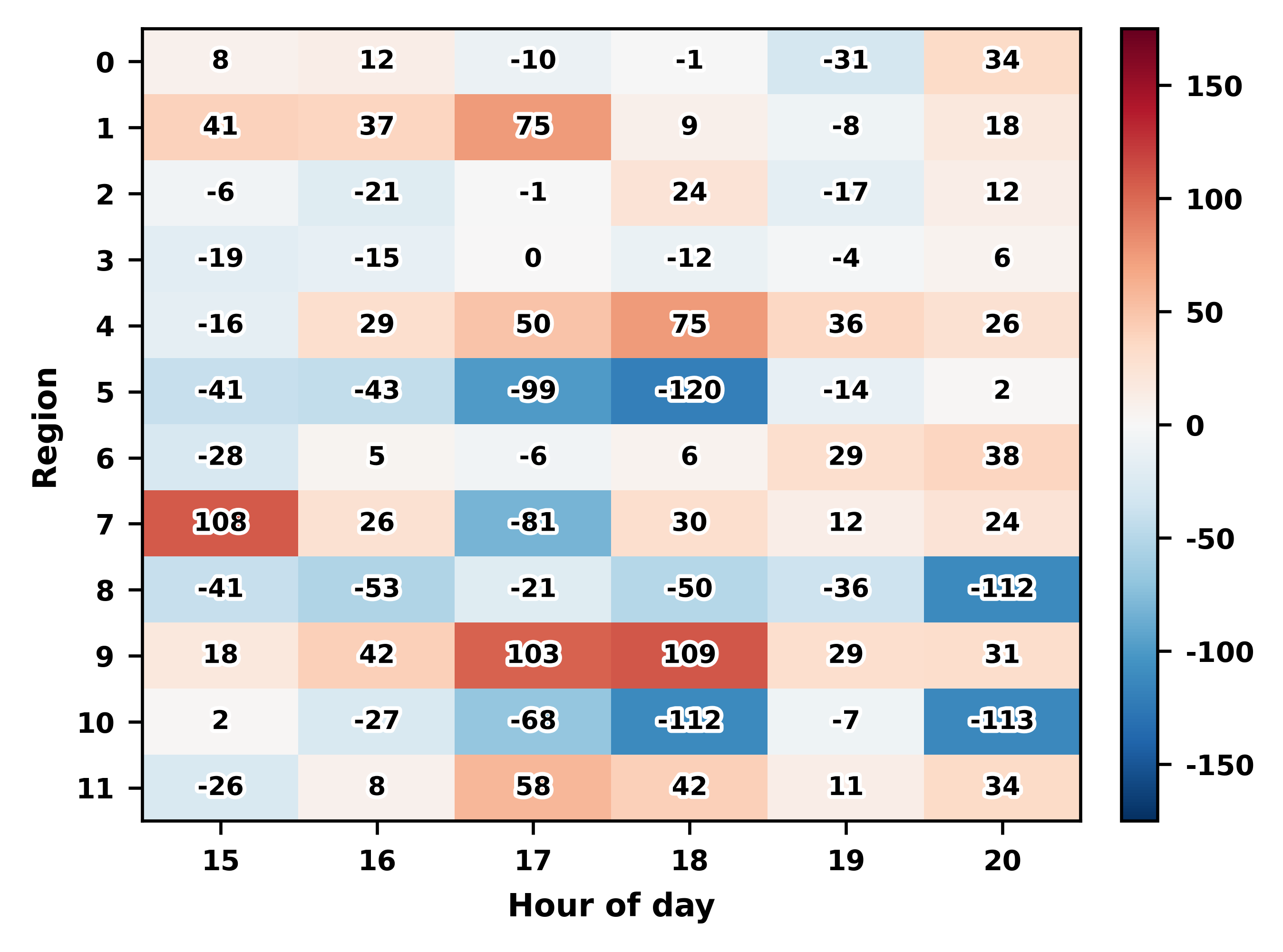}%
    }\hfill
    \subfloat[Mean wait time (min), unregulated\label{fig:surge_wait_unreg}]{%
        \includegraphics[width=0.32\linewidth]{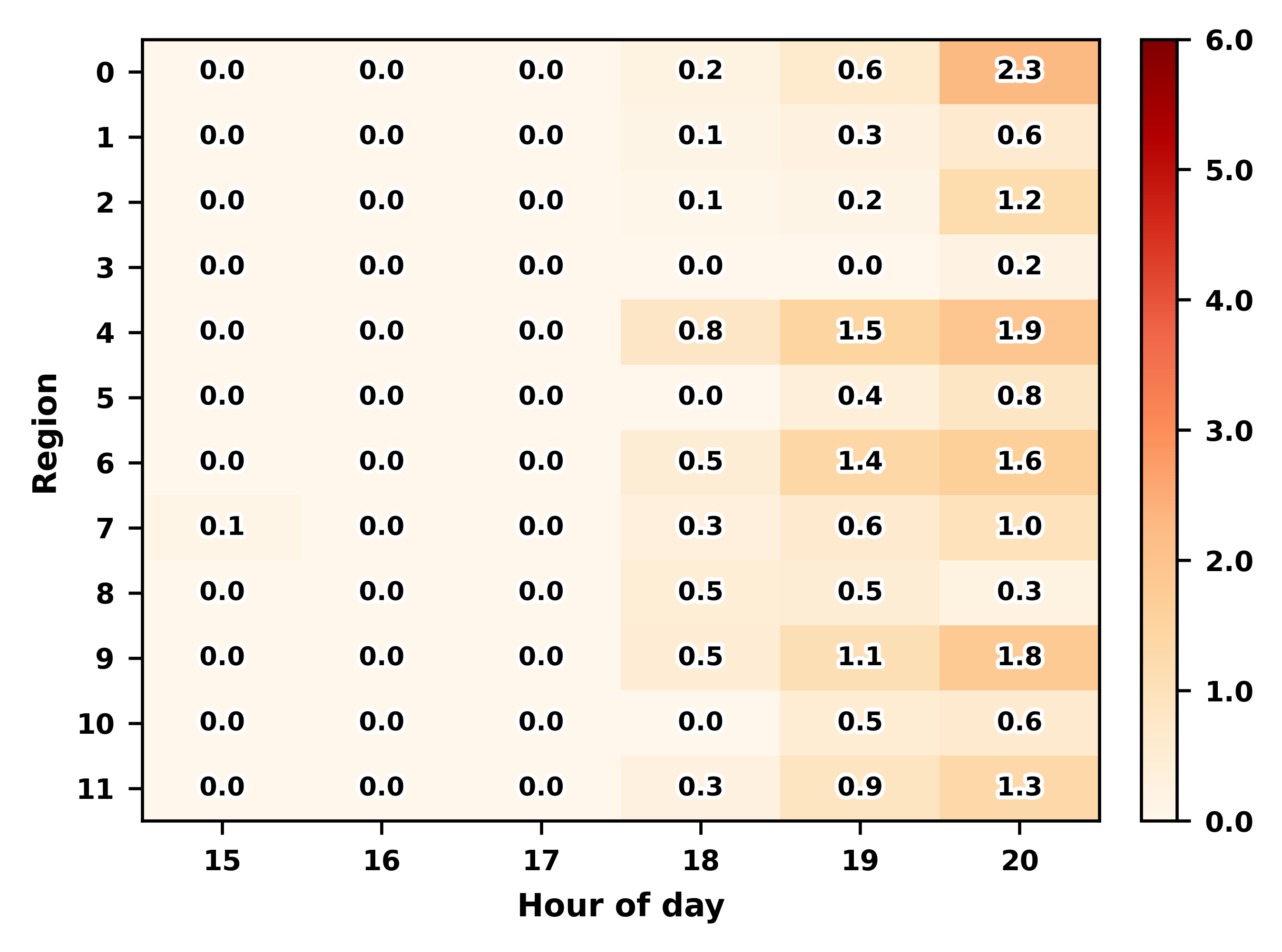}%
    }\hfill
    \subfloat[Lost AMoD demand (\#), unregulated\label{fig:surge_unserved_unreg}]{%
        \includegraphics[width=0.32\linewidth]{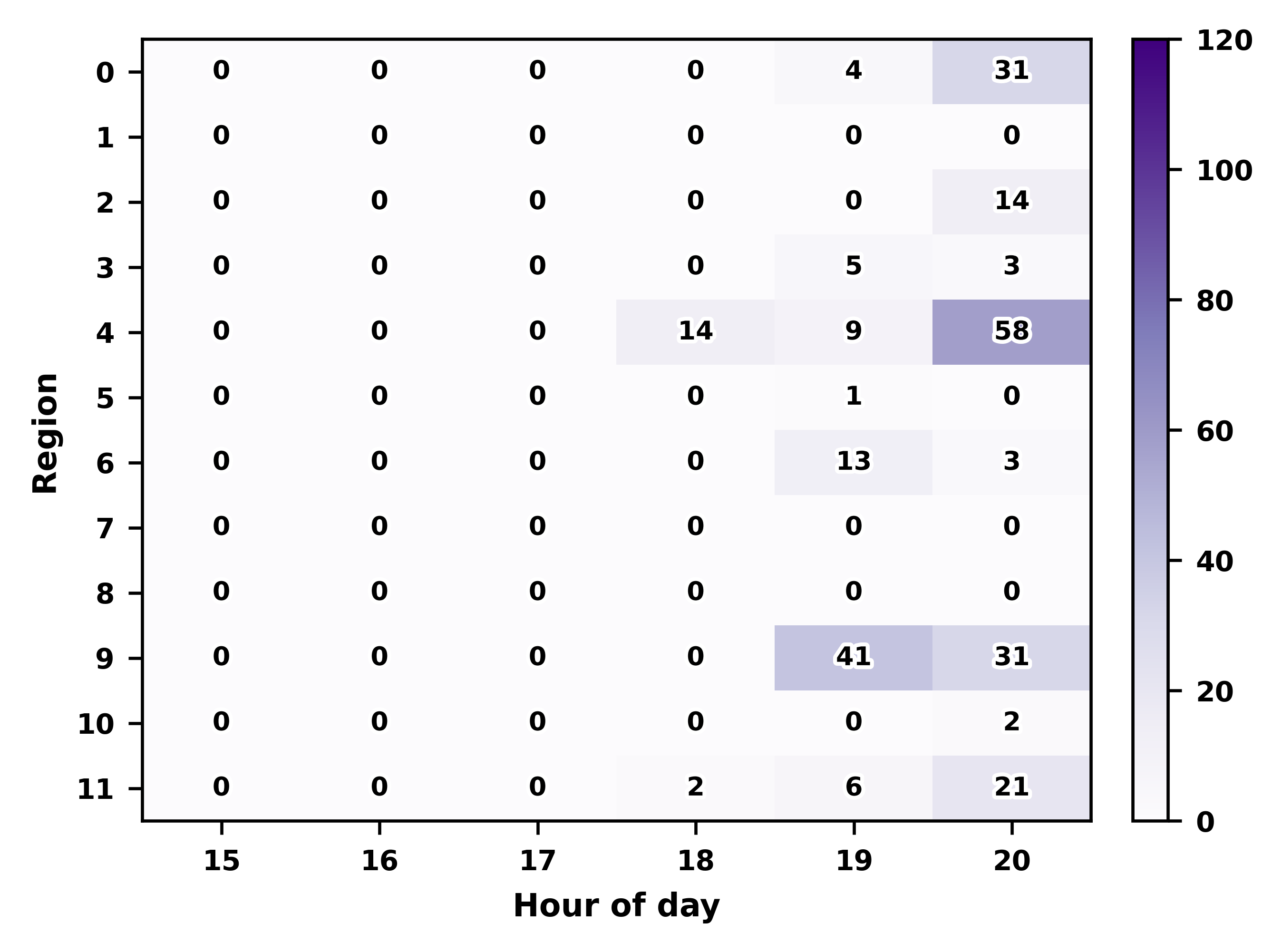}%
    }
    \par\vspace{-1em}
    \subfloat[Rebalancing net flow (\#), regulated\label{fig:surge_reb_reg}]{%
        \includegraphics[width=0.32\linewidth]{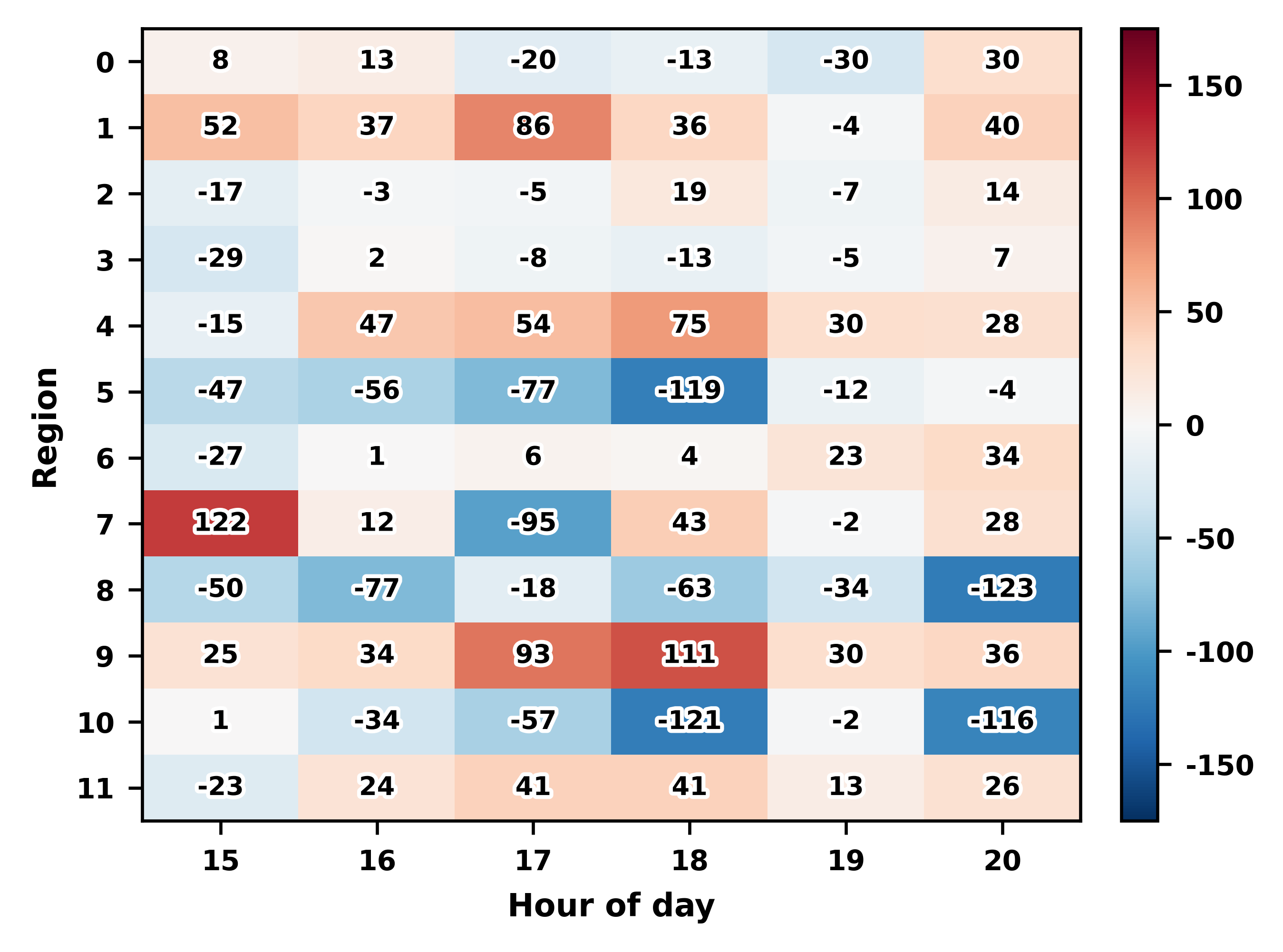}%
    }\hfill
    \subfloat[Mean wait time (min), regulated\label{fig:surge_wait_reg}]{%
        \includegraphics[width=0.32\linewidth]{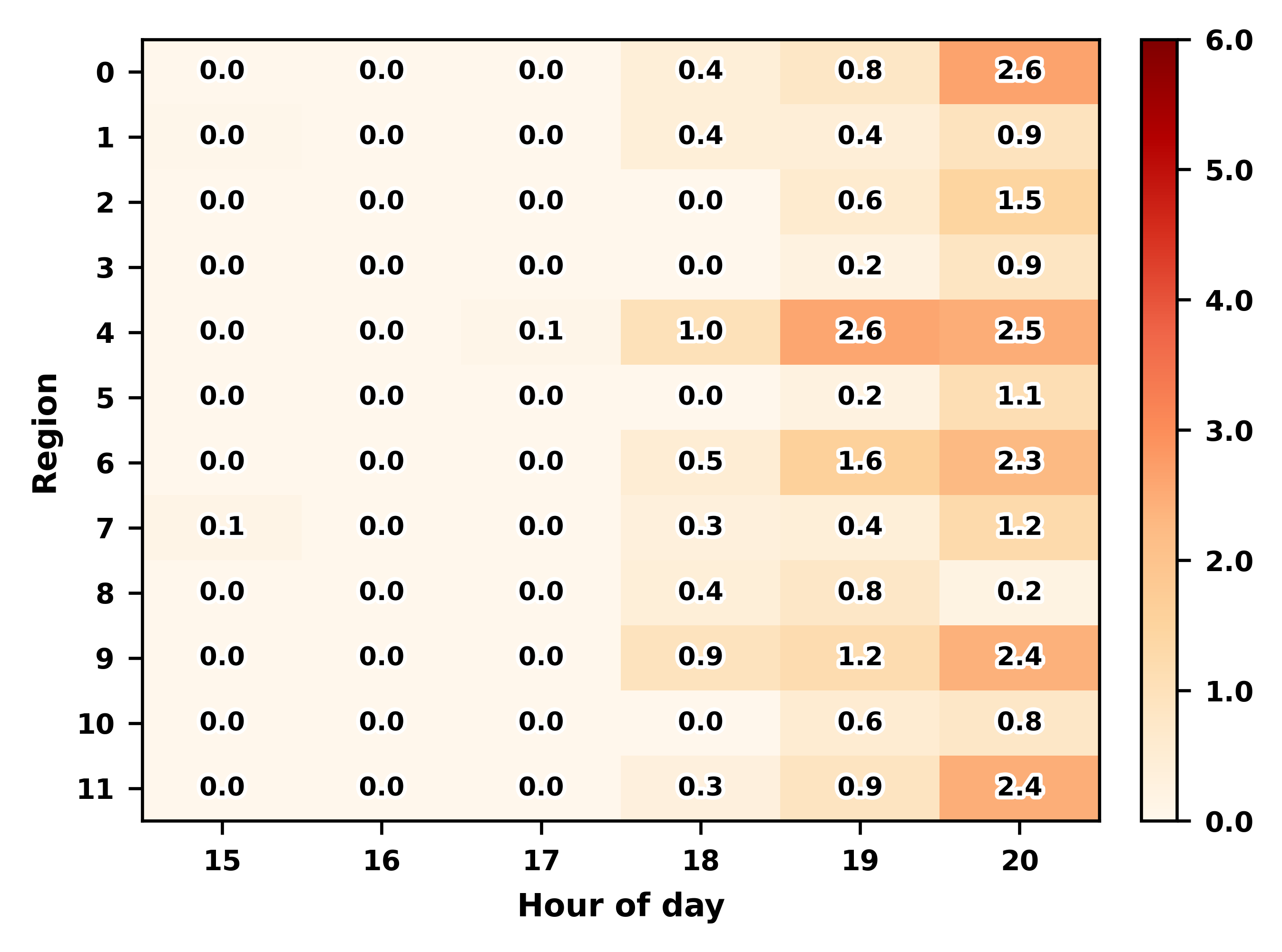}%
    }\hfill
    \subfloat[Lost AMoD demand (\#), regulated\label{fig:surge_unserved_reg}]{%
        \includegraphics[width=0.32\linewidth]{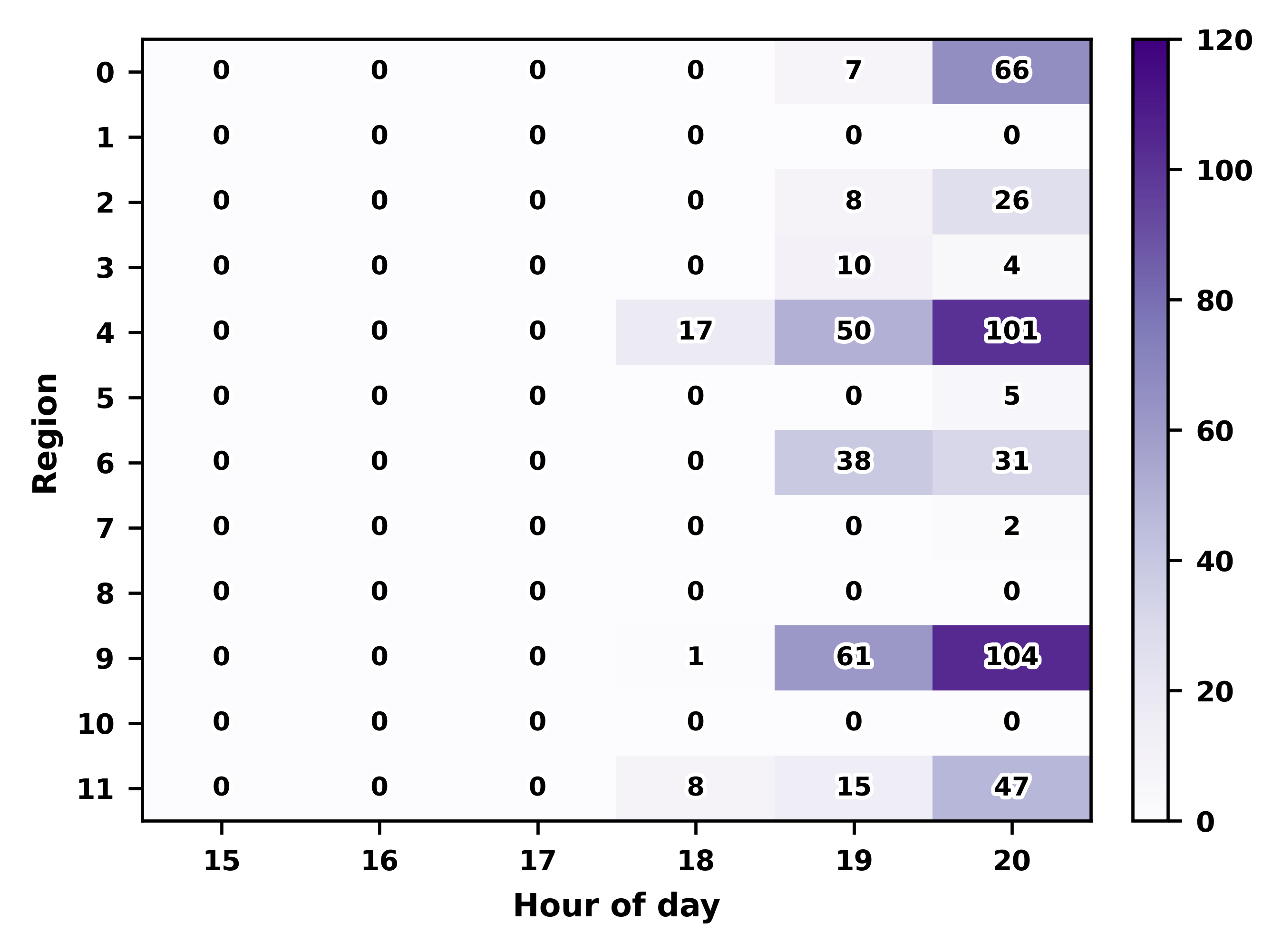}%
    }

    \caption[Rebalancing policy and service quality under static surge pricing regulation]{Rebalancing policy and service quality under static surge pricing regulation by region and hour in one representative evaluation run.}
    \label{fig:surge_policy_response}
\end{figure}

\end{minipage}
\end{strip}

\clearpage

\section{Hourly Vehicle Utilization under Regional Service Availability Regulation}
\label{app:util_spatial}

\begin{table}[!ht]
\centering
\caption[Hourly vehicle utilization under spatial equity regulation]{Vehicle utilization by hour in percent under regional service availability regulation.}
\label{tab:spatial_utilization_hourly}
\footnotesize
\begin{tabular}{llrrrrrr}
\toprule
\textbf{Regulation}
& \textbf{} 
& \multicolumn{6}{c}{\textbf{Hour of day}} \\
\cmidrule(lr){3-8}
& 
& \textbf{15} 
& \textbf{16} 
& \textbf{17} 
& \textbf{18} 
& \textbf{19} 
& \textbf{20} \\
\midrule
Static
& $\phi^\mathrm{SA}=0$  
& 37 & 28 & 42 & 83 & 96 & 97 \\

\cmidrule(lr){1-8}

\multirow[t]{4}{*}{Dynamic}
& $\lambda=1.0$  
& 41 & 29 & 43 & 84 & 97 & 98 \\
& $\lambda=0.1$  
& 41 & 33 & 48 & 85 & 97 & 98 \\
& $\lambda=0.05$  
& 42 & 35 & 48 & 87 & 97 & 98 \\
& $\lambda=0.01$  
& 43 & 36 & 50 & 86 & 97 & 98 \\

\cmidrule(lr){1-8}

Static
& $\phi^\mathrm{SA}=200$ 
& 43 & 36 & 49 & 86 & 97 & 97 \\
\bottomrule
\end{tabular}

\vspace{0.5em}
\begin{minipage}{\linewidth}
\footnotesize
\centering
\textit{Notes:} Values are shown for one representative evaluation run.
\end{minipage}
\end{table}


\bibliographystyle{IEEEtran}
\bibliography{literature}